\documentclass[lettersize,journal]{IEEEtran}
\usepackage{amsmath,amsfonts,amsthm,amssymb}
\newtheorem{remark}{Remark}
\usepackage{algorithmic}
\usepackage{algorithm}
\usepackage{array}
\usepackage[caption=false,font=normalsize,labelfont=sf,textfont=sf]{subfig}
\usepackage{textcomp}
\usepackage{stfloats}
\usepackage{url}
\usepackage{verbatim}
\usepackage{graphicx}
\usepackage{xcolor}
\usepackage{bm}
\usepackage{caption}
\usepackage{subcaption}
\usepackage{booktabs}
\usepackage{cite}
\usepackage{tabularx}
\usepackage{tikz}
\usetikzlibrary{arrows.meta,calc}

\DeclareMathOperator*{\argmin}{arg\,min}
\DeclareMathOperator*{\argmax}{arg\,max}

\begin{document}

\title{3D-Force Estimation for Dual-arm Manipulation of Deformable Linear Objects via Factor Graphs}

\author{
Annika Delucchi,
Vincenzo Di~Paola,
Bruno Marco Quadrelli,
Matteo Zoppi
\thanks{A. Delucchi, V. Di~Paola and M. Zoppi are with the University of Genova, Department of Mechanical, Energy, Management and Transportation Engineering, Via all'opera Pia 15A, Genova, Italy.}
\thanks{M. B. Quadrelli is with the Jet Propulsion Laboratory, California Institute of Technology, 4800 Oak Grove Drive, Pasadena, United States.}
\thanks{Corresponding author: annika.delucchi@edu.unige.it.}
}

\maketitle

\begin{abstract}
Estimating forces acting on Deformable Linear Objects (DLOs) is challenging due to their underactuated nature. 
However, reliable contact estimation is crucial to precisely control them and enable safe manipulation tasks.
Differently from the various approaches presented in the literature, this work adopts a probabilistic formulation where the DLO state and sensors data are related through conditional distributions. In particular, factor graphs are used to determine magnitude, direction and location of a 3D force acting on a DLO, while simultaneously reconstructing its shape. 
Since DLOs are not actuated and typically thin -- like cables and wires -- strain gauges and actuation readings are usually unpractical. Therefore, to overcome this issue, this work is tested using one \textit{type} of sensing information only thus developing a generic technique suited for a wider variety of tasks. 
Realistic tests are made by introducing noise on the measurements while the DLO follows a prescrived path in presence of external forces accurately estimated. 
\end{abstract}


\section{Introduction}  
\IEEEPARstart{D}{}ual-arm manipulation of Deformable Linear Objects refers to the class of tasks where a slender flexible object --- like ropes and cables --- is guided into a desired shape by imposing a sequence of poses at its end tips. Indeed, threading, knotting and routing~\cite{CuiralZueco2024,Caporali2024,Renda2026DualManip,Delucchi2026} are just a few of the recurring movements necessary for their application in healthcare domain~\cite{BurgnerKahrs2015, Dupont2022, Webster2006,Orekhov2015}, search and rescue~\cite{TognonGabellieri2018}, space applications~\cite{Russo2023Space} and so on. Beyond their impressive motion range, a primary motivation in studying DLOs lies in their intrinsic compliance making them well-suited for contact-rich tasks. In this context, although the physics of contact has been extensively studied in the past under both modelling and computational perpectives~\cite{Kikuchi1994,Konyukhov2013,Beal2003,Chamekh2009, GayNeto2015,Tasora2020}, only recently researchers considered the contact in the field of DLOs' manipulation~\cite{Delucchi2026Eucomes,Ferguson2026dlo}. The explanation for this delay must be searched in the underactuated nature of DLOs which makes the contact estimation and thus their control highly complex. Therefore, we employ factor graphs~\cite{DellaertBook-FactorGraphs} to foresee the contact along the DLO thus addressing one of the limitations that prevent their application in the aforementioned tasks.

Though actuated and then different\footnote{The DLO model does not include any actuation term.}, this approach has been applied in the proximal field of continuum robots which will provide the main reference for this work. In this context, early research focused on estimating loads applied at the robot's tip~\cite{Xu2016,Wang2023Survey}. 
As the field further evolved, several approaches were proposed to estimate external forces acting also along robot's backbone~\cite{Lilge2023,Diezinger2022,Pourafzal2023}, combining vision-based information with other sensors, like actuation readings~\cite{FeliuTalegon2025}, strain gauges~\cite{Ferguson2024, Wang2026CosimoRendaLaschi}, Fiber Bragg Grating~\cite{Lilge2025} and laser scans~\cite{FergusonRuckerWebster2024}.
Among them, it is possible to distinguish two main directions. The first formulates force estimation as an inverse problem, exploiting the kinematic or mechanical model of the robot and using sensors to infer the unknown external forces. Instead, the second adopts a probabilistic formulation, where the robot's state and sensor measurements are related through conditional probability distributions, allowing force estimation even in weakly observable scenarios.



%

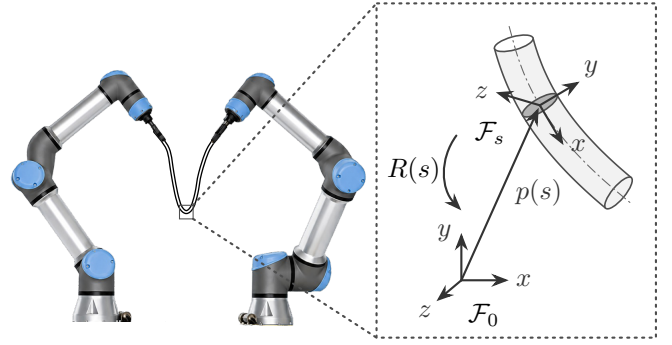
\begin{figure}[tb]
    \centering
    \begin{tikzpicture}[>=Stealth, line cap=round]
    \def\kr{0.62}    
    \def\ki{0.66}    
    \pgfmathsetmacro{\rw}{3.3*\kr}   

    \begin{scope}[scale=\kr]
        \coordinate (SL) at (3.10,4.07);
        \coordinate (L2) at (3.37,3.70);
        \coordinate (SR) at (4.48,4.05);
        \coordinate (R2) at (4.23,3.68);
        \coordinate (Cm) at (3.765,2.35);   

        \draw[draw=black, double=white, double distance=0.6pt, line width=0.5pt]
            plot[smooth, tension=0.6] coordinates {
                (SL) (L2) (3.47,3.18) (3.55,2.75) (3.64,2.46) (Cm)
                (3.89,2.46) (3.98,2.75) (4.06,3.18) (R2) (SR)};

        \node[anchor=south west, inner sep=0] at (0,0)
            {\includegraphics[width=\rw cm]{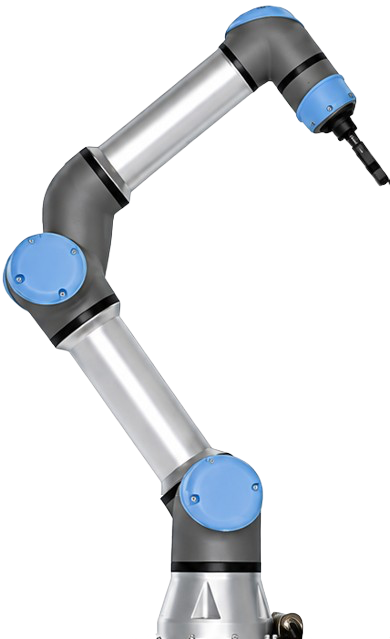}};
        \node[anchor=south west, inner sep=0] at (4.29,-0.08)
            {\includegraphics[width=\rw cm]{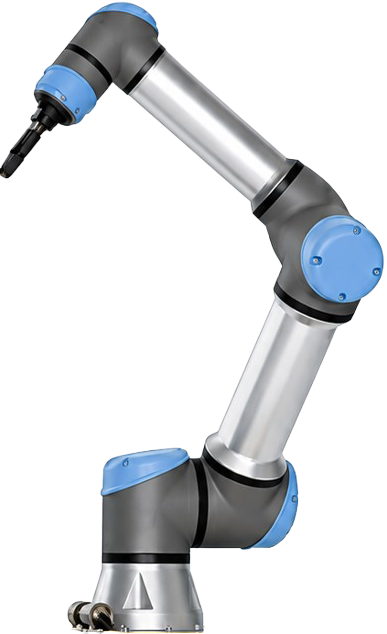}};
    \end{scope}

    \begin{scope}[shift={(5.61,0.39)}, scale=\ki]
        \draw[dotted, thick, black!70] (-1.15,-0.9) rectangle (4.45,5.8);
        \coordinate (BTL) at (-1.15,5.8);
        \coordinate (BBL) at (-1.15,-0.9);
        \fill[black!5] (1.986,5.105) -- (2.010,5.018) -- (2.036,4.932) -- (2.064,4.847) -- (2.092,4.763) -- (2.123,4.680) -- (2.154,4.598) -- (2.187,4.516) -- (2.222,4.435) -- (2.259,4.355) -- (2.297,4.275) -- (2.337,4.196) -- (2.378,4.118) -- (2.422,4.039) -- (2.467,3.961) -- (2.514,3.884) -- (2.564,3.806) -- (2.615,3.729) -- (2.668,3.652) -- (2.724,3.574) -- (2.782,3.497) -- (2.842,3.420) -- (2.905,3.342) -- (2.970,3.265) -- (3.037,3.187) -- (3.107,3.109) -- (3.180,3.031) -- (3.255,2.952) -- (3.332,2.873) -- (3.413,2.794) -- (3.496,2.714) -- (3.582,2.634) -- (3.671,2.553) -- (3.763,2.472) -- (3.858,2.390) -- (3.956,2.307) -- (3.444,1.693) -- (3.339,1.781) -- (3.237,1.869) -- (3.138,1.957) -- (3.041,2.045) -- (2.947,2.133) -- (2.856,2.220) -- (2.767,2.308) -- (2.681,2.395) -- (2.597,2.483) -- (2.516,2.570) -- (2.437,2.658) -- (2.361,2.746) -- (2.287,2.834) -- (2.215,2.923) -- (2.146,3.011) -- (2.080,3.100) -- (2.015,3.190) -- (1.953,3.280) -- (1.893,3.370) -- (1.835,3.461) -- (1.779,3.552) -- (1.726,3.644) -- (1.675,3.736) -- (1.626,3.829) -- (1.579,3.923) -- (1.534,4.017) -- (1.490,4.112) -- (1.449,4.207) -- (1.410,4.303) -- (1.373,4.400) -- (1.338,4.497) -- (1.304,4.596) -- (1.272,4.695) -- (1.242,4.794) -- (1.214,4.895) -- cycle;
        \draw[black!70, semithick] plot[smooth] coordinates {(1.986,5.105) (2.010,5.018) (2.036,4.932) (2.064,4.847) (2.092,4.763) (2.123,4.680) (2.154,4.598) (2.187,4.516) (2.222,4.435) (2.259,4.355) (2.297,4.275) (2.337,4.196) (2.378,4.118) (2.422,4.039) (2.467,3.961) (2.514,3.884) (2.564,3.806) (2.615,3.729) (2.668,3.652) (2.724,3.574) (2.782,3.497) (2.842,3.420) (2.905,3.342) (2.970,3.265) (3.037,3.187) (3.107,3.109) (3.180,3.031) (3.255,2.952) (3.332,2.873) (3.413,2.794) (3.496,2.714) (3.582,2.634) (3.671,2.553) (3.763,2.472) (3.858,2.390) (3.956,2.307)};
        \draw[black!70, semithick] plot[smooth] coordinates {(1.214,4.895) (1.242,4.794) (1.272,4.695) (1.304,4.596) (1.338,4.497) (1.373,4.400) (1.410,4.303) (1.449,4.207) (1.490,4.112) (1.534,4.017) (1.579,3.923) (1.626,3.829) (1.675,3.736) (1.726,3.644) (1.779,3.552) (1.835,3.461) (1.893,3.370) (1.953,3.280) (2.015,3.190) (2.080,3.100) (2.146,3.011) (2.215,2.923) (2.287,2.834) (2.361,2.746) (2.437,2.658) (2.516,2.570) (2.597,2.483) (2.681,2.395) (2.767,2.308) (2.856,2.220) (2.947,2.133) (3.041,2.045) (3.138,1.957) (3.237,1.869) (3.339,1.781) (3.444,1.693)};
        \draw[black!70, semithick, fill=white, rotate around={15.26:(1.600,5.000)}] (1.600,5.000) ellipse (0.400 and 0.132);
        \draw[black!70, semithick, fill=black!5, rotate around={50.19:(3.700,2.000)}] (3.700,2.000) ellipse (0.400 and 0.132);
        \draw[black!60, dash pattern=on 4pt off 1.5pt on 0.6pt off 1.5pt, thin]
            (1.508,5.338) -- plot[smooth] coordinates {(1.600,5.000) (1.626,4.906) (1.654,4.813) (1.684,4.721) (1.715,4.630) (1.748,4.540) (1.782,4.450) (1.818,4.362) (1.856,4.273) (1.896,4.186) (1.938,4.099) (1.981,4.013) (2.026,3.927) (2.074,3.842) (2.123,3.757) (2.175,3.672) (2.228,3.588) (2.284,3.504) (2.342,3.421) (2.402,3.337) (2.464,3.254) (2.529,3.171) (2.596,3.088) (2.665,3.005) (2.737,2.923) (2.811,2.840) (2.888,2.757) (2.968,2.674) (3.050,2.590) (3.134,2.507) (3.222,2.423) (3.312,2.339) (3.404,2.255) (3.500,2.171) (3.599,2.085) (3.700,2.000)} -- (3.969,1.776);


        \def\ls{1.5}   
        \coordinate (S) at (2.123,3.757);
        \draw[black!70, semithick, fill=black!30, rotate around={30.76:(S)}] (S) ellipse (0.400 and 0.132);
        \draw[->, thick, black!80] (S) -- ($(S)+\ls*(0.317,-0.533)$)  node[right=-1pt]       {$x$};
        \draw[->, thick, black!80] (S) -- ($(S)+\ls*(0.533,0.317)$)   node[above right=-2pt] {$y$};
        \draw[->, thick, black!80] (S) -- ($(S)+\ls*(-0.589,0.174)$)  node[left=-1pt]        {$z$};

        \node at ($(S)+(-1.00,-0.55)$) {$\mathcal{F}_s$};   

        \coordinate (W) at (0.550,0.250);
        \draw[->, thick, black!80] (W) -- ++(0,0.95)  node[left=0pt]  {$y$};
        \draw[->, thick, black!80] (W) -- ++(0.95,0)  node[right=-1pt] {$x$};
        \draw[->, thick, black!80] (W) -- ++(-0.5,-0.42) node[below left=-2pt] {$z$};
        \draw[-{Stealth[length=2.4mm]}, thick, black!80, shorten >=1pt] (W) -- node[pos=0.5, right=2pt] {$p(s)$} (S);
        \draw[->, thick, black!80] (0.45,3.15) to[bend right=40] (0.55,1.65);
        \node at (-0.40,2.45) {$R(s)$};
        \node at (1,-0.45) {$\mathcal{F}_0$};   

    \end{scope}

    \draw[thin, black!70] ($(Cm)+(-0.09,-0.09)$) rectangle ($(Cm)+(0.09,0.09)$);
    \draw[dotted, thick, black!70] ($(Cm)+(0.09,0.09)$) -- (BTL);
    \draw[dotted, thick, black!70] ($(Cm)+(0.09,-0.09)$) -- (BBL);
    \end{tikzpicture}
    \caption{Left: two manipulators hold the ends of a DLO to change its shape. Right (zoomed view): the DLO is modelled as a generic rod undergoing large displacements and small deformations. Each cross-section, at arc length $s$, is associated with a reference frame $\mathcal{F}_s$ whose $x$-axis is tangent to the centreline. The cross-section position $p(s)$ and orientation $R(s)$ are expressed with respect to the world frame $\mathcal{F}_0$.}
    \label{fig:rod-cont-scheme}
\end{figure}

For the mentioned reasons, this article leverages the Graph-based method for the estimation of 3D forces acting on DLOs. Hence, without any assumption, location, direction and magnitude of the contact force are estimated simultaneously from discrete spatial information only. During the estimation, only discrete positions along the rod are assumed to be available through camera tracking, emulating a motion capture system based on reflective markers. This minimal sensing setup is dictated by the will of a practical solution, while existing approaches typically combine multiple sensing modalities~\cite{Caporali2024,Hu2025}. The use of a single \textit{type} of information further emphasizes the challenge of estimating contact forces from limited measurements while simultaneously reconstructing the rod configuration.

The article is structured as follows. Section~\ref{sec:kirchhoff-model} reviews the Kirchhoff inextensible model, focusing on the proposed application. Section~\ref{sec:factor-graph} introduces factor graphs and details the components for setting up the estimation problem. Section~\ref{sec:simulation} reports simulation results and Section~\ref{sec:conclusion} summarises and concludes the work.

\section{Kirchhoff Inextensible Model}
\label{sec:kirchhoff-model}
The behaviour of slender, inextensible rods can be modelled using Kirchhoff inextensible theory~\cite{Cosserat1909} and Lie groups~\cite{Selig2004}.

Indicating with $L$ the rod length and with $s\in[0,L]$ the curvilinear coordinate along the rod axis, the rod spatial evolution is described by the motion of its rigid cross-section (Fig.~\ref{fig:rod-cont-scheme}). At every $s$, the cross-section is endowed with a mobile frame $\mathcal{F}_s$, whose $x-$axis is aligned with the rod centreline. Cross-sectional poses $T(s)$ are expressed with respect to a world frame $\mathcal{F}_0$ and represented as elements of $SE(3)$,
\begin{align*}
T(s) = \begin{pmatrix}
    R(s) & p(s) \\ 0 & 1
\end{pmatrix},   
\end{align*}
with $R(s) \in SO(3)$ rotation matrix and $p(s) \in \mathbb{R}^{3}$ position vector. Thus, the rod configuration space is given by $G := \{ T | s \in [0,L] \rightarrow T(s) \in SE(3)\}$.

The rod's space variation $\frac{d}{ds}(\cdot) = (\cdot)'$ can be described as
\begin{align}
\label{eq:dTds}
    &T(s)' = T(s)\hat{\xi}(s), \\
    &\hat{\xi}(s) = \begin{pmatrix}
        \tilde{\kappa}(s) & \rho(s) \\ 0 & 0
    \end{pmatrix}. \notag
\end{align}
Here $\xi(s)$ is an element of the Lie algebra $se(3)$, with $\kappa(s)$ and $\rho(s)$ angular and linear strain, respectively; $\hat{(\cdot)}: \mathbb{R}^6 \rightarrow \mathbb{R}^{4\times4}$ is the hat operator and $\tilde{(\cdot)}:\mathbb{R}^3 \rightarrow \mathbb{R}^{3\times3}$ the skew operator. While other models let $\rho$ assume arbitrary values, the Kirchhoff inextensible one establishes constant value $\rho =~ \begin{pmatrix}L & 0 & 0\end{pmatrix}$, representing fixed elongation and no shear.

The constitutive equation that relates the strain $\xi(s)$ to the stress $\Lambda(s) \in se^*(3)$ is given as
\begin{align*}
    \Lambda(s) = K(\xi(s) - \xi^*(s),
\end{align*}
where $\xi^*(s)$ is the reference or undeformed strain and $K(s) = \operatorname{diag}(G J_x, E I_y, E I_z, E A, G A, G A)$ is the stiffness matrix containing the material's Young's $E$ and shear $G$ moduli, the rod cross-sectional area $A$, the polar moment of inertia $J_x$ and the principal planar area moments of inertia $I_y, I_z$. In general, $K(s)$ depends on $s$, but the homogeneous material assumption makes it constant along the rod.

Recalling Euler-Poincaré equations for statics, the equilibrium equation reads
\begin{align*}
    \Lambda(s)' = ad_{\xi}^\top \Lambda(s) - w_d,
\end{align*}
with $w_d$ distributed loads and adjoint operator $\operatorname{ad}: se(3) \rightarrow se(3)$,
\begin{align*}
    \operatorname{ad}_{\xi} = \begin{pmatrix}
        \tilde{\kappa} & 0 \\ \tilde{\rho} & \tilde{\kappa}
    \end{pmatrix}.
\end{align*}
Finally, the rod shape is obtained solving the system of equations
\begin{align*}
    \begin{cases}
         T(s)' &= T(s)\hat{\xi}(s) \\
         \Lambda(s) &= K(\xi(s) - \xi^*(s) \\
         \Lambda(s)' &= ad_{\xi}^\top \Lambda(s) - w_d.
    \end{cases}
\end{align*}

\subsection{Local Geometric Variable Strain}
\label{sec:local-gvs}
The strain vector $\xi(s)$ of a Kirchhoff inextensible rod can be approximated with base functions as
\begin{align*}
    &\xi(s) = 
    \begin{pmatrix}
        B\,\Phi(s)\,q  \\ e_1
    \end{pmatrix},
\end{align*}
where $B$ is a selection matrix enabling only the angular strain components, $\Phi(s)$ is the matrix of shape functions, $q$ is a vector of generalised coordinates and $e_1 = (1\,\, 0\,\, 0)^\top$ enforces the inextensibility and no-shear assumptions.
Recalling~\cite{Boyer2019IEEE}, the kinematics equations are forward integrated as in Eq.~\eqref{eq:dTds} with initial conditions $T(0) = I$ and parametrizing the orientation $R(s)$ with unit quaternions.
Then, given the terminal conditions $\Lambda(1) = 0$ and $Q(1) = 0$, the following equations are backward integrated 
\begin{align}
    \begin{cases}
        \Lambda' = \operatorname{ad}_{\xi(s)}^\top \Lambda(s) - w_d \\
        Q' = (B\,\Phi(s))^\top \, \Lambda(s),
    \end{cases}
    \label{eq:dLambda, dQ}
\end{align}
to obtain the stress resultant $\Lambda(s)$ and the contribution of the generalised external loads $Q(s)$. The static equilibrium is enforced as
\begin{align}
    K_e \, q + Q(q,0) - J(q,s_c)^\top w_c(s_c) -  J(q,1)^\top \Lambda(1) = 0,
    \label{eq:equil-constraint}
\end{align}
where $w_d$ and $w_c$ denote distributed and concentrated loads, respectively, $\Lambda(1)$ is the wrench imposing the constraint pose at the rod end and $K_e$ is the generalised stiffness matrix with dimensions compatible with $q$.

For dual-manipulation, the forward kinematics Eq.~\eqref{eq:dTds} and the backward statics Eq.~\eqref{eq:dLambda, dQ} are solved as a Boundary Value Problem for the equilibrium constraint Eq.~\eqref{eq:equil-constraint} with $T(0) = I$ and $T(1) = T_{des}$, being $T_{des}$ the relative pose between the rod tips imposed by the manipulation.

Apart from the different polynomial bases that one can use, it is known that local bases embody concentrated forces better than global ones~\cite{Mathew2024}. Therefore, throughout this work we use a local bases made of linear Lagrange polynomials. 

\section{Factor Graphs and Force Estimation} 
\label{sec:factor-graph}
Factor graphs are a family of probabilistic graphical models particularly suited to solve inference problems~\cite{DellaertBook-FactorGraphs}.
They are bipartite graphs $\mathcal{F} = (\mathcal{U}, \mathcal{V}, \mathcal{E})$ consisting of two types of nodes, variables  $x_j \in \mathcal{V}$ and factors $\phi_i \in \mathcal{U}$, connected by edges $e_{ij} \in \mathcal{E}$.    
Here, variables $x_j$ represent the state and are subject to estimation; their values are found solving the optimization problem. Instead, factors $\phi_i$ contain information about the system; a graph can include different types of factors, each of them associated with a density and a cost term. Even if densities strongly depend on the kind of sensor, it is common to use multi-variate Gaussian distribution with probability density
\begin{align*}
    \mathcal{N}(\theta;\mu,\Sigma) = \frac{1}{\sqrt{2\pi\Sigma}}\,
    e^{\textstyle\big(-\frac{1}{2}\,\lVert\theta-\mu\rVert_{\Sigma}^{2}\big)},
\end{align*}
 with $\theta$ unknown variables, $\mu$ mean, $\Sigma$ covariance matrix and $||\theta-\mu ||_\Sigma^2 = (\theta-\mu)^\top \Sigma^{-1}(\theta-\mu)$ defined as the squared Mahalanobis distance~\cite{Mahalanobis1936}. In many cases it is both reasonable and convenient to model the factor noise as corrupted by zero-mean Gaussian noise, yielding
 \begin{align*}
     || e_i||^2 = ||\theta-\mu ||_\Sigma^2 = e_i^\top \Sigma_i^{-1}  e_i,
 \end{align*}
 from which descends the Gaussian potential
\begin{align}
     \phi_i \propto e^{\textstyle\big(-\frac{1}{2}\,\lVert e_i\rVert^{2}\big)}.
     \label{eq:factor-gauss-pot}
\end{align}
 Unlike other probabilistic methods, factors graphs allow to specify a joint density as a product of factors
 \begin{align*}
     \phi(X) = \prod_i \phi_i(X_i),
 \end{align*}
 and for an arbitrary factor graph, the Maximum a Posteriori (MAP) estimation corresponds to maximizing the factor product
 \begin{align}
     X^{MAP} = \argmax_X \phi(X) = \argmax_X  \prod_i \phi_i(X_i).
     \label{eq:factors-product}
 \end{align}
Here, $X_i$ contains the variable nodes $x_j$ related to the factor $\phi_i$, and $X$ all the system's state variables. 
Assuming that all factors are in the form of Gaussian potentials as Eq.~\eqref{eq:factor-gauss-pot}, thus implying that likelihood factors are derived from Gaussian noise, the estimation can be cast into a non-linear least squares optimization. Applying the negative logarithm to Eq.~\eqref{eq:factors-product} transforms the product of exponential Gaussian potentials into a sum of squared residuals~\cite[\S\,2.2]{DellaertBook-FactorGraphs}, yielding
 \begin{align}
     X^{MAP} = \argmin_X J(X),
     \label{eq:opt_problem}
 \end{align}
 where $X$ contains the unknown state variables $x_j$
 and the cost function $J(X)=\frac{1}{2}\sum_i ||e_i||^2$ collects the cost of each factor $\phi_i$ in the graph.

\subsection{Rod Discretization}
Through the use of factor graphs, the rod shape reconstruction problem is solved for a finite set of nodes along its centreline. Consistent with this framework, the continuous rod behaviour is discretized into $N$ variable nodes.
The kinematics at each node reads
\begin{align*}
    T_{i+1} = T_i\, \exp \,(\xi_i \, \Delta s),
\end{align*}
with $\Delta s = s_{i+1} - s_i$ the interval between the nodes and $\exp: se(3) \rightarrow SE(3)$ exponential map. The
piecewise constant approximation of the strain is
\begin{align}
    \xi_i = \frac{1}{2} K^{-1}(\Lambda_i + \Lambda_{i+1}) + \xi^* + n_{i,\xi}.
    \label{eq:xi-linear}
\end{align}
Here, $n_{i,\xi} \sim \mathcal{N}(0, \Sigma_\xi)$ encodes the modelling errors due to discretization and linear approximation.

The discretized equilibrium accounts also for the unknown contact force that we want to estimate. More specifically, 
assuming no external moments acting on the rod and adopting the external force parametrization of~\cite{Aditya2026}, the effect of punctual forces is modelled as that of distributed loads and parametrized with as many Gaussian bases functions as the external forces
\begin{align*}
    f_{ext} = \sum_{h=1}^M \begin{pmatrix}
        \alpha_{x,h} \\ \alpha_{y,h} \\ \alpha_{z,h}
    \end{pmatrix}
    \frac{1}{\sqrt{2\pi}\sigma_h} e^{\Big(-\frac{(s-\mu_h)^2}{2\sigma_h^2}\Big)}.
\end{align*}
Thus, the force estimation problem reduces to find:
$\mu_h$, representing the Gaussian mean that corresponds to the force location; $\sigma_i$, the standard deviation indicating the width of the Gaussian; $\alpha_h = \begin{pmatrix} \alpha_{x,h} & \alpha_{y,h} & \alpha_{z,h} \end{pmatrix}^\top$, the Gaussian amplitude and force magnitude.
For concentrated contact forces, $\sigma_h$ can be set to a fixed value representing point loads. Also, to guarantee that the estimated $\mu_h$ remains within the rod domain $[0,L]$, the following bounded parametrization is used
\begin{align*}
    \mu_h(\beta_h) =\frac{L}{2} (\tanh(\beta_h) +1).
\end{align*}
The force location can thus be expressed in terms of $\beta_h$, which maps $\mu_h$ into the arc-length interval $[0,L]$. Consequently, the vector of the estimation variables is defined as $\theta = [\alpha_1^\top, \beta_1, \dots, \alpha_M^\top, \beta_M] \in \mathbb{R}^{4\times M}$. Following~\cite{Aditya2026}, the cumulative force is defined as
\begin{align*}
    v_{ext,i} &= \int_0^{s_i} f_{ext}(x) dx \\
     &= \small{\frac{1}{2} \sum_{h=1}^M \begin{pmatrix}
        \alpha_{x,h} \\ \alpha_{y,h} \\ \alpha_{z,h} \end{pmatrix} \Big[ 
     \operatorname{erf}\Big(-\frac{s_i-\mu_h}{\sqrt{2}\sigma_h}\Big) - \operatorname{erf}\Big(-\frac{-\mu_h}{\sqrt{2}\sigma_h}\Big) \Big]}.
\end{align*}
As a consequence, over the interval $[s_i, s_i+1]$ the discretized equilibrium equation updates to
\begin{align}
    &\Lambda_{i+1} =  \operatorname{Ad}(T_{i}^{-1}T_{i+1})^\top\Lambda_i
       - \operatorname{Ad}(T_{i+1})^\top\begin{pmatrix} 0 \\ w_g +f_{ext} \end{pmatrix} + n_\Lambda\notag \\ 
    &= \scriptsize{\operatorname{Ad}(T_{i}^{-1}T_{i+1})^\top\Lambda_i 
    - \operatorname{Ad}(T_{i+1})^\top\begin{pmatrix}  0 \\ w_g + (v_{ext,i+1} - v_{ext,i}) \end{pmatrix} + n_\Lambda.}
    \label{eq:equilib-discrete}
\end{align}
Here, $w_g = \begin{pmatrix} 0 & 0 & \rho g A
\end{pmatrix}^\top$ stands for the gravity vector and $n_\Lambda$ accounts for the error due to moment approximation.

\subsection{Problem Definition}
To solve the simultaneous force and rod shape estimation, the problem is structured as follows.

The rod is discretized into $N$ variables nodes equally spaced, corresponding to $N-1$ segments of length $\Delta s$. Two of the nodes coincide with the rod tips. Each of these nodes is associated with two unknowns, the pose $T_i$ and the internal wrench $\Lambda_i$: these are the estimated quantities that determine the rod shape.
To bind their value to a physical solution for dual-manipulation, some factors have to be enforced. Tip poses $\bar{T}_{base}, \bar{T}_{tip}$ are given as prior factors; 
kinematics consistency and equilibrium between the nodes are instead ensured through dedicated mechanics factors. 
During the manipulation, the rod is subject to an unknown external force $f_{ext}$ at an unknown location $s$. This force must be estimated and included in the equilibrium equation to effectively reconstruct the rod shape. Being subject to estimation, also the force is represented as a variable node associated with its own prior, and linked to the graph through equilibrium factors. Shape data are available to provide a ground-truth for the rod shape. In particular, $N-2$ positions along the rod are known at a distance $\Delta s$ from one another. 

\begin{figure*}
    \centering
    \includegraphics[scale=1]{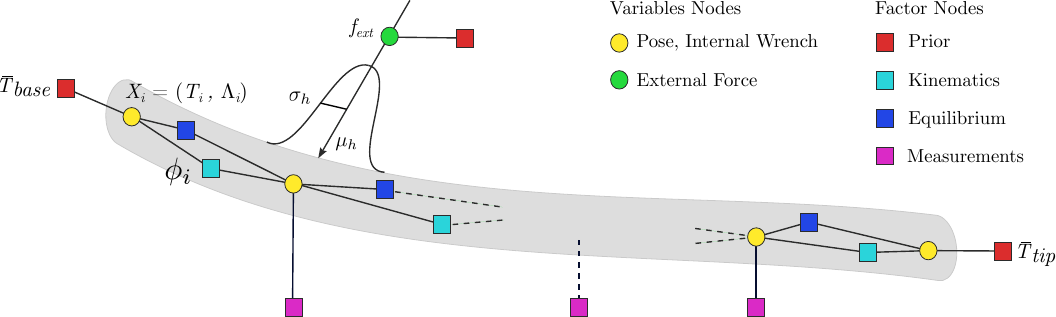}
    \caption{Graphical representation of the factor graph used to solve the force and shape estimation problem in superimposition over a gray rod. Squares indicate factors nodes, while variable nodes are circles. Red squares represent prior factors, blue ones equilibrium, cyan ones kinematics and magenta ones are sensors measurements. Yellow nodes encode the rod state, accounting for pose and internal wrench, while the green one represents force.  }
    \label{fig:graph}
\end{figure*}

A graphical representation of the problem is provided in Fig.~\ref{fig:graph} and the following section details the residuals and costs associated to each factor in the graph.


\subsection{Variable and Factors Nodes}
Recalling the previous section, the rod state consists of poses $T_i$ and internal wrenches $\Lambda_i$ encoded into $N$ variables nodes
\begin{align*}
    X_i = (T_i, \Lambda_i), \hspace{1cm} i = 1, \dots, N.
\end{align*}
These nodes are connected by factors nodes, providing physical grounding and sensors observations. 
Notice that each factor connects only to the variable nodes of which it is a function.
To encode prior, mechanics and sensors knowledge, different factors are employed.


\subsubsection{Prior Factors}
represent known data with known uncertainty that must be satisfied within the graph solution. In the dual manipulation setting, the rod tip poses $\bar{T}_{base}, \bar{T}_{tip}$ are enforced through prior factors on the first and last nodes. The associated residual vectors read
\begin{align*}
    e_1 = \log(\bar{T}_{base}^{-1} \, T_1)^\vee, \hspace{0.7cm}
    e_N = \log(\bar{T}_{tip}^{-1} \, T_N)^\vee,
\end{align*}
that, using the vee operator $(\cdot)^\vee : \mathbb{R}^{4\times4}\rightarrow \mathbb{R}^{6} $, yield the cost terms
\begin{align*}
    J_1 = \frac{1}{2} e_1^\top \, \Sigma_0^{-1} \, e_1, \hspace{0.8cm}
    J_N = \frac{1}{2} e_N^\top \, \Sigma_0^{-1} \, e_N.
\end{align*}
The covariance $\Sigma_0$ is the same for both priors. 
\subsubsection{Mechanics Factors}
ensure physics consistency between the rod nodes. In this work we distinguish between kinematics and equilibrium factors. 
Recalling Eq.~\eqref{eq:xi-linear}, \textit{Kinematics factors} exploit the definition of strain between two adjacent nodes, assuming constant strain $\xi$ over the interval $\Delta s$ between the nodes. 
Hence, the residual over this factor reads
\begin{align*}
    e_{i,\xi} = \frac{1}{2} K^{-1}(\Lambda_i + \Lambda_{i+1}) + \xi^* - \frac{1}{\Delta s}\,\log\,(T_i^{-1}\,T_{i+1}),
\end{align*}
with cost term
   $J_{i,\xi} = \frac{1}{2} e_{i,\xi}^\top \, \Sigma_\xi^{-1} \, e_{i,\xi}$.

Instead, \textit{Equilibrium Factors} impose equilibrium between adjacent nodes considering the wrench balance Eq.~\eqref{eq:equilib-discrete} in the residual form
\begin{align*}
    e_{i,\Lambda} &= \Lambda_{i+1} -  \operatorname{Ad}(T_{i}^{-1}T_{i+1})^\top\Lambda_i + \operatorname{Ad}(T_{i+1})^\top\begin{pmatrix} 0 \\ w_g +f_{ext} \end{pmatrix}  
\end{align*}
with cost term 
    $J_{i,\Lambda} = \frac{1}{2} e_{i,\Lambda}^\top \, \Sigma_{\Lambda}^{-1} \, e_{i,\Lambda}$.
\subsubsection{Measurements Factors} encapsulate sensor data, providing a ground-truth from real-world. Including data from external readings helps addressing the ill-conditioned nature of the problem. Measurement data can origin from a broad variety of sensors. In this work, we assume to know only positions at discrete points along the rod, coming from a motion capture system. For simplicity, we assume to know $N-2$ positions, corresponding to the position vector $p_i$ of each node. 
The residual for each node reads
\begin{align*}
    e_{i,p} = p_{i,graph} - p_{i,meas},
\end{align*}
with cost $J_{p} = \frac{1}{2}e_{i,p}^\top \, \Sigma_p^{-1} \, e_{i,p}$ and ${\Sigma_p}$ indicating the sensor uncertainty.
\subsection{Cost Function}
At this point, it is possible to explicit the cost function $J(X)$. The graph contains priors factor on both tips, kinematics and equilibrium factors on each node and measurements factors if available. Therefore, the optimization problem in Eq.~\eqref{eq:opt_problem} has the cost function
\begin{align*}
     J(X) &= J_{prior}(X) +  J_{mech}(X) +  J_{meas}(X) \\
     & =  J_1(X_1) + J_N(X_N)  + \sum_{i = 1}^N J_{i,\xi}(X_i) \, + \\&+\sum_{i = 1}^N J_{i,\Lambda}(X_i) + \sum_{i = 2}^{N-1} J_{p}(X_i).
\end{align*}

\section{Simulations}
\label{sec:simulation}
To evaluate the effectiveness of the proposed approach for dual-arm manipulation, the factor graph method is implemented in C++ using the GTSAM library~\cite{gtsam} and solved with the Levenberg-Marquard algorithm. To this end, a $1$\,m long Nitinol rod was considered, with circular cross-section of $1.5\times10^{-3}$\,m diameter, Young's modulus $E=50$\,GPa and shear modulus $G=30$\,GPa. 
The graph is constructed representing the rod with $N=20$ 
nodes, two of them coinciding with the rod tips.

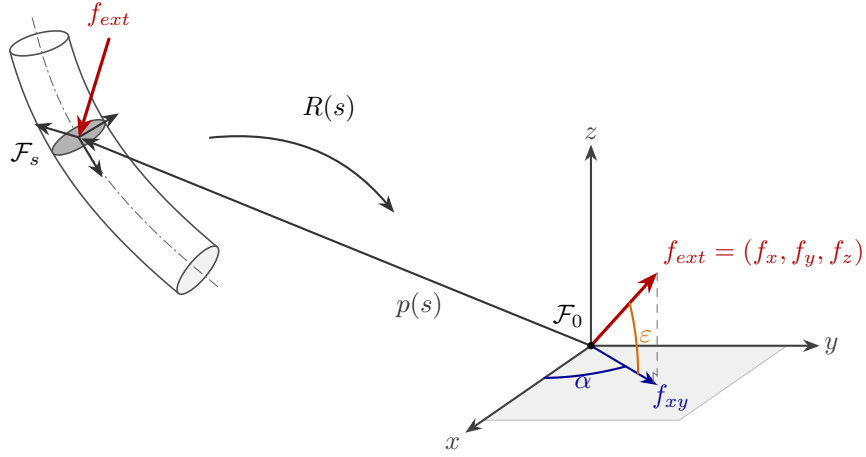
\begin{figure*}[tb]
    \centering
    \begin{tikzpicture}[>=Stealth, line cap=round]
    \begin{scope}
    \fill[black!5] (1.986,5.105) (2.010,5.018) (2.036,4.932) (2.064,4.847) (2.092,4.763) (2.123,4.680) (2.154,4.598) (2.187,4.516) (2.222,4.435) (2.259,4.355) (2.297,4.275) (2.337,4.196) (2.378,4.118) (2.422,4.039) (2.467,3.961) (2.514,3.884) (2.564,3.806) (2.615,3.729) (2.668,3.652) (2.724,3.574) (2.782,3.497) (2.842,3.420) (2.905,3.342) (2.970,3.265) (3.037,3.187) (3.107,3.109) (3.180,3.031) (3.255,2.952) (3.332,2.873) (3.413,2.794) (3.496,2.714) (3.582,2.634) (3.671,2.553) (3.763,2.472) (3.858,2.390) (3.956,2.307) -- (3.444,1.693) (3.339,1.781) (3.237,1.869) (3.138,1.957) (3.041,2.045) (2.947,2.133) (2.856,2.220) (2.767,2.308) (2.681,2.395) (2.597,2.483) (2.516,2.570) (2.437,2.658) (2.361,2.746) (2.287,2.834) (2.215,2.923) (2.146,3.011) (2.080,3.100) (2.015,3.190) (1.953,3.280) (1.893,3.370) (1.835,3.461) (1.779,3.552) (1.726,3.644) (1.675,3.736) (1.626,3.829) (1.579,3.923) (1.534,4.017) (1.490,4.112) (1.449,4.207) (1.410,4.303) (1.373,4.400) (1.338,4.497) (1.304,4.596) (1.272,4.695) (1.242,4.794) (1.214,4.895) -- cycle;
    \draw[black!70, semithick] plot[smooth] coordinates {(1.986,5.105) (2.010,5.018) (2.036,4.932) (2.064,4.847) (2.092,4.763) (2.123,4.680) (2.154,4.598) (2.187,4.516) (2.222,4.435) (2.259,4.355) (2.297,4.275) (2.337,4.196) (2.378,4.118) (2.422,4.039) (2.467,3.961) (2.514,3.884) (2.564,3.806) (2.615,3.729) (2.668,3.652) (2.724,3.574) (2.782,3.497) (2.842,3.420) (2.905,3.342) (2.970,3.265) (3.037,3.187) (3.107,3.109) (3.180,3.031) (3.255,2.952) (3.332,2.873) (3.413,2.794) (3.496,2.714) (3.582,2.634) (3.671,2.553) (3.763,2.472) (3.858,2.390) (3.956,2.307)};
    \draw[black!70, semithick] plot[smooth] coordinates {(1.214,4.895) (1.242,4.794) (1.272,4.695) (1.304,4.596) (1.338,4.497) (1.373,4.400) (1.410,4.303) (1.449,4.207) (1.490,4.112) (1.534,4.017) (1.579,3.923) (1.626,3.829) (1.675,3.736) (1.726,3.644) (1.779,3.552) (1.835,3.461) (1.893,3.370) (1.953,3.280) (2.015,3.190) (2.080,3.100) (2.146,3.011) (2.215,2.923) (2.287,2.834) (2.361,2.746) (2.437,2.658) (2.516,2.570) (2.597,2.483) (2.681,2.395) (2.767,2.308) (2.856,2.220) (2.947,2.133) (3.041,2.045) (3.138,1.957) (3.237,1.869) (3.339,1.781) (3.444,1.693)};
    \draw[black!70, semithick, fill=white, rotate around={15.26:(1.600,5.000)}] (1.600,5.000) ellipse (0.400 and 0.132);
    \draw[black!70, semithick, fill=black!5, rotate around={50.19:(3.700,2.000)}] (3.700,2.000) ellipse (0.400 and 0.132);
    \draw[black!60, dash pattern=on 4pt off 1.5pt on 0.6pt off 1.5pt, thin]
        (1.508,5.338) -- plot[smooth] coordinates {(1.600,5.000) (1.626,4.906) (1.654,4.813) (1.684,4.721) (1.715,4.630) (1.748,4.540) (1.782,4.450) (1.818,4.362) (1.856,4.273) (1.896,4.186) (1.938,4.099) (1.981,4.013) (2.026,3.927) (2.074,3.842) (2.123,3.757) (2.175,3.672) (2.228,3.588) (2.284,3.504) (2.342,3.421) (2.402,3.337) (2.464,3.254) (2.529,3.171) (2.596,3.088) (2.665,3.005) (2.737,2.923) (2.811,2.840) (2.888,2.757) (2.968,2.674) (3.050,2.590) (3.134,2.507) (3.222,2.423) (3.312,2.339) (3.404,2.255) (3.500,2.171) (3.599,2.085) (3.700,2.000)} -- (3.969,1.776);
    
    \coordinate (S) at (2.123,3.757);
    \draw[black!70, semithick, fill=black!30, rotate around={30.76:(S)}] (S) ellipse (0.400 and 0.132);
    \draw[->, thick, black!80] (S) -- ++(0.317,-0.533);
    \draw[->, thick, black!80] (S) -- ++(0.533,0.317);
    \draw[->, thick, black!80] (S) -- ++(-0.589,0.174);
    
    \node at ($(S)+(-0.70,-0.22)$) {$\mathcal{F}_s$};   
    \draw[-{Stealth[length=2.8mm]}, very thick, red!70!black] ($(S)+(0.40,1.30)$) -- (S)
        node[pos=0, above=1pt, red!70!black] {$f_{ext}$};
    \end{scope}

    \begin{scope}[shift={(8.9,1.0)}, %
        x={(-0.396cm,-0.274cm)}, y={(0.72cm,0cm)}, z={(0cm,0.72cm)}, 
        axis/.style={-{Stealth[length=2.2mm]}, thick, black!75},
        guide/.style={dashed, black!55, thin},
        forcevec/.style={-{Stealth[length=2.8mm]}, very thick, red!70!black},
        projvec/.style={-{Stealth[length=2.2mm]}, thick, blue!60!black},
        ang/.style={thick, blue!60!black},
        angE/.style={thick, orange!85!black},
        every node/.style={inner sep=1.5pt}]
    \def\az{50}      
    \def\el{35}      
    \def\len{3.6}    
    \pgfmathsetmacro{\fx}{\len*cos(\el)*cos(\az)}
    \pgfmathsetmacro{\fy}{\len*cos(\el)*sin(\az)}
    \pgfmathsetmacro{\fz}{\len*sin(\el)}

    \coordinate (O) at (0,0,0);

    \fill[black!6] (0,0,0) -- (3.6,0,0) -- (3.6,3.6,0) -- (0,3.6,0) -- cycle;
    \draw[black!25] (0,0,0) -- (3.6,0,0) -- (3.6,3.6,0) -- (0,3.6,0) -- cycle;

    \draw[axis] (0,0,0) -- (4.2,0,0) node[below left]  {$x$};
    \draw[axis] (0,0,0) -- (0,4.2,0) node[right]       {$y$};
    \draw[axis] (0,0,0) -- (0,0,3.7) node[above]       {$z$};

    \draw[guide] (\fx,\fy,\fz) -- (\fx,\fy,0);
    \draw[projvec] (0,0,0) -- (\fx,\fy,0)
        node[pos=0.85, below right, blue!60!black] {$f_{xy}$};
    \draw[forcevec] (0,0,0) -- (\fx,\fy,\fz)
        node[above right, red!70!black] {$f_{ext}=(f_x,f_y,f_z)$};

    \draw[ang] plot[domain=0:\az, samples=30, variable=\t]
        ({1.55*cos(\t)}, {1.55*sin(\t)}, 0);
    \node[blue!60!black] at ({1.95*cos(\az/2)}, {1.95*sin(\az/2)}, 0) {$\alpha$};

    \draw[angE] plot[domain=0:\el, samples=30, variable=\t]
        ({2.1*cos(\t)*cos(\az)}, {2.1*cos(\t)*sin(\az)}, {2.1*sin(\t)});
    \node[orange!85!black] at
        ({2.55*cos(\el/2)*cos(\az)}, {2.55*cos(\el/2)*sin(\az)}, {2.55*sin(\el/2)}) {$\varepsilon$};

    \draw[black!55, thin]
        ({\fx-0.18*cos(\az)},{\fy-0.18*sin(\az)},0) --
        ({\fx-0.18*cos(\az)},{\fy-0.18*sin(\az)},0.18) --
        (\fx,\fy,0.18);
    \fill (0,0,0) circle (1.3pt);
    \end{scope}

    \draw[-{Stealth[length=2.4mm]}, thick, black!80, shorten >=1pt] (O) -- (S)
        node[pos=0.28, below left=-2pt] {$p(s)$};
    \draw[->, thick, black!80]
        ($(S)!0.22!(O)!0.65cm!90:(O)$) to[bend left=28] ($(S)!0.58!(O)!0.65cm!90:(O)$);
    \node at ($(S)!0.40!(O)!1.6cm!90:(O)$) {$R(s)$};
    \fill (O) circle (1.3pt);
    \node at ($(O)+(-0.30,0.40)$) {$\mathcal{F}_0$};
    \end{tikzpicture}
    \caption{The force $f_{ext}$ applied at a location $s$ along the rod and expressed in the local frame $\mathcal{F}_s$ is rotated in the world frame $\mathcal{F}_0$ to extract its direction. The force direction is indicated in terms of two angles $\alpha$ and $\varepsilon$, representing the azimuth angle and the elevation, respectively.}
    \label{fig:az-el}
\end{figure*}

\subsection{Statistical Analysis}
To asses the accuracy of the estimation, we considered two trajectories that excite the bending and torsion modes of the DLO through generic forces acting on it. This procedure proves the generality of the proposed method while considering the worst-case scenario. These tests will therefore aim at estimating the magnitude, direction and the location of the external force by using deflection information only. As mentioned, the \textit{type} of information that we use to infer the strains is unique: there is no sensor fusion or mix of measurement types as done in many preceding works for TDCR.  
As a reference, in~\cite{FeliuTalegon2025} authors estimate 1-D and 3-D external forces acting on tendon-driven continuum robots using actuation readings while assuming to know the force location. Similarly,~\cite{Wang2026CosimoRendaLaschi} presents a 3-D force estimator based on embedded stretch sensors, while explicitly assuming prior knowledges of the force location. Furthermore, the approaches~\cite{Ferguson2024, Aloi2019} impose simplifying assumptions on the force orientation, like constraining forces to local cross-sectional planes.

Here, we select only one sensor \textit{type} to stress even more  the robustness of the method.

\begin{remark}
Although probabilistic approaches were used in the field of continuum robots, literature comparison cannot be effectively done as the physical systems are sensibly different. The continuum manipulators have stiffer structure than the sole DLO. This motivates the following considerations on the force parameters which are only demonstrative and do not limit the generality of the approach. 
\end{remark}

\subsubsection{Dataset Generation}
\label{sec:dataset-gen}
The measurement data -- or dataset -- are generated by introducing noise in the solution of the model in Sec.~\ref{sec:local-gvs}, to emulate rod measures.
In particular, 3D forces are applied at generic locations $s$ and the shape is reconstructed by solving a Boundary Value Problem for given end tip poses  $\bar{T}_{base},\bar{T}_{tip}$.
From this \textit{baseline} solution, tip poses are inserted into the graph as priors, while $9$ equally spaced positions along the rod are extracted and given as position measurements to the estimation problem, emulating markers placed at $0.1$m from each other. 

\subsubsection{Measurements Noise}
Measurement values are corrupted by additive zero-mean white Gaussian noise to model optical tracking conditions, which typically achieve sub-millimetre accuracy under optimal setups. To systematically evaluate the performance of the estimator across realistic tracking conditions, we apply 
the noise $n_{sensor} \sim \mathcal{N}(0, 1\times10^{-4}\text{ m})$ to the position measurements as $p_{i,pos} = p_{i,pos} + n_{sensor}$. 
This corresponds to a maximum 3D offset of $5\times10^{-4}$\,m from the true marker position.

\subsubsection{Force Parameters Range}
Differently from the actuated robots, here we are forced to limit the possible range of magnitude that the external force can take. Indeed, small forces cause big variation of the DLO's configuration and, on the other hand, small forces would be undetectable as they do not produce sensible shape changes. Therefore, we set the maximum force magnitude to be $1.5$ N for non-axial components and $0.3$ N for the axial one\footnote{These values can be adapted based on the task and DLO material and geometry without loss of generality.}. The axial effect of the force is also considered but, as the section is small, its magnitude remains limited. 

Regarding the region of contact, this is set to occur in the central part of the DLO $s\in[0.3,\,0.7]$ as, practically, the manipulators' end-effectors must be free to move during manipulation tasks.

\subsubsection{Initialization}
The estimation problem is initialised with initial straight rod, $T_i = I, \Lambda_i = 0 \, \, \forall \, \, i = 1,\dots N$. The force variables for a single force $M=1$ are $\theta = [\alpha_x, \alpha_y, \alpha_z, \beta]$ and are initialised to zero force $\theta = [10^{-4}, 10^{-4}, 10^{-4}, 0]$. Observe that $\beta=0$ implies $s=\mu=0.5$, hence the initial guess location is at the middle of the rod. Instead, the standard deviation of the Gaussian parametrizing the force is set to $\sigma_f = 1.2\times10^{-2}$\,m, representing the span of concentrated forces.

\subsubsection{Covariances}
Solving the estimation problem also requires selecting the values for the factors' covariances. Here, they are selected as 
\begin{align*}
    \Sigma_0 &= \operatorname{diag}(0.01, \, 0.01, \, 0.01, \, 0.001, \, 0.001, \, 0.001)  &\text{[rad,m]},\\
    \Sigma_\xi &= \operatorname{diag}(0.15, \, 0.15, \, 0.15, \, 0.001, \, 0.001, \, 0.001)   &\text{[rad/m,-]}, \\
    \Sigma_\Lambda &= \operatorname{diag}(0.0001, \, 0.0001, \, 0.0001, \, 0.1, \, 0.1, \, 0.1)   &\text{[N/m]}, \\
    \Sigma_p &= \operatorname{diag}(0.0001, \, 0.0001, \, 0.0001)  &\text{[m]}, \\
    \Sigma_{f} &= \operatorname{diag}(2, \, 2, \, 2, \, 3)  &\text{[N,-]}.
\end{align*}
All noise models are assumed to be diagonal and isotropic, therefore with uncorrelated components and no preferred direction.

\begin{figure*}[p]
    \centering
    \begin{minipage}[t]{0.68\textwidth}
        \centering\offinterlineskip
        \includegraphics[width=0.45\linewidth]{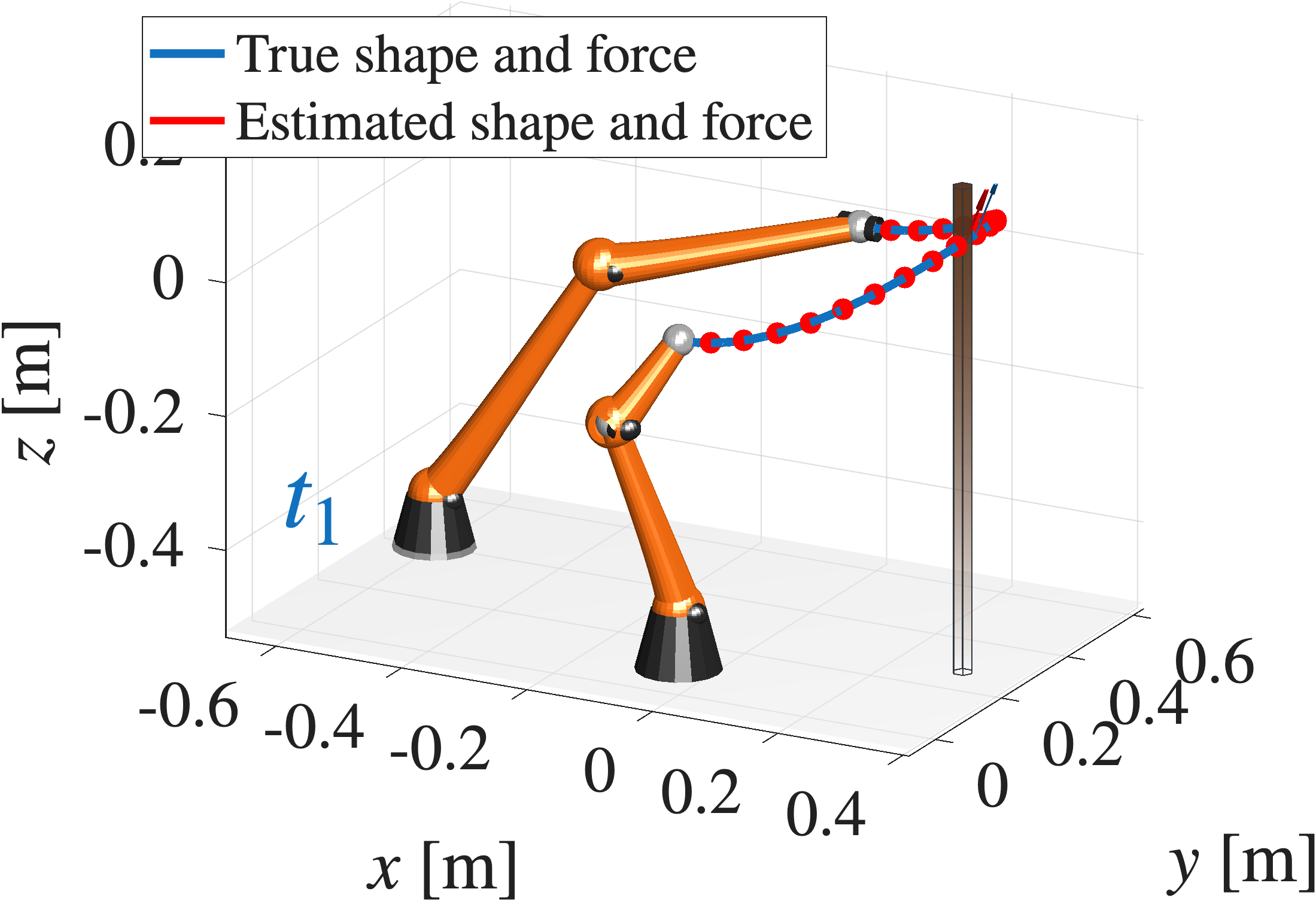}\hfill
        \includegraphics[width=0.45\linewidth]{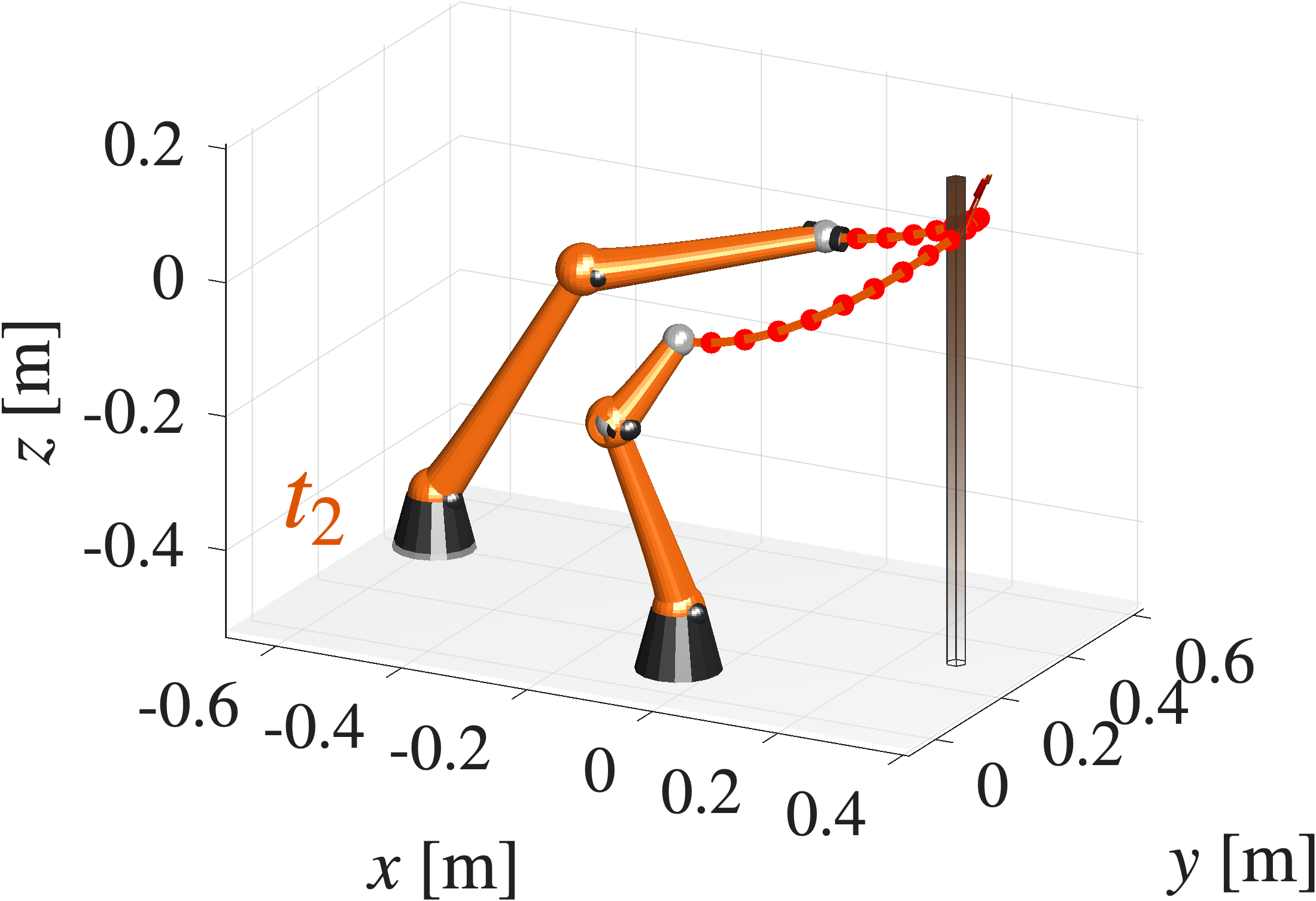}\par\vspace{1mm}
        \includegraphics[width=0.45\linewidth]{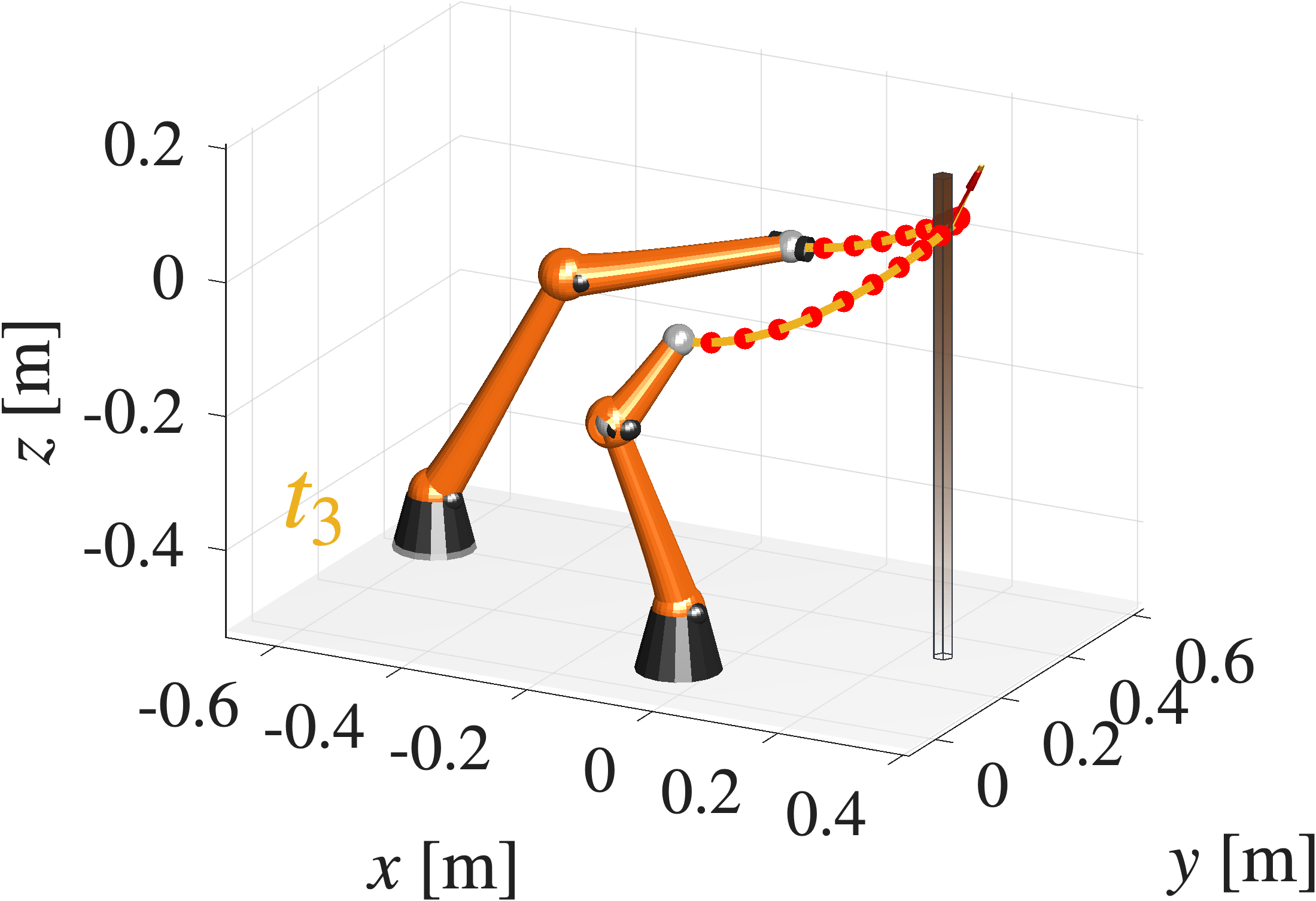}\hfill
        \includegraphics[width=0.45\linewidth]{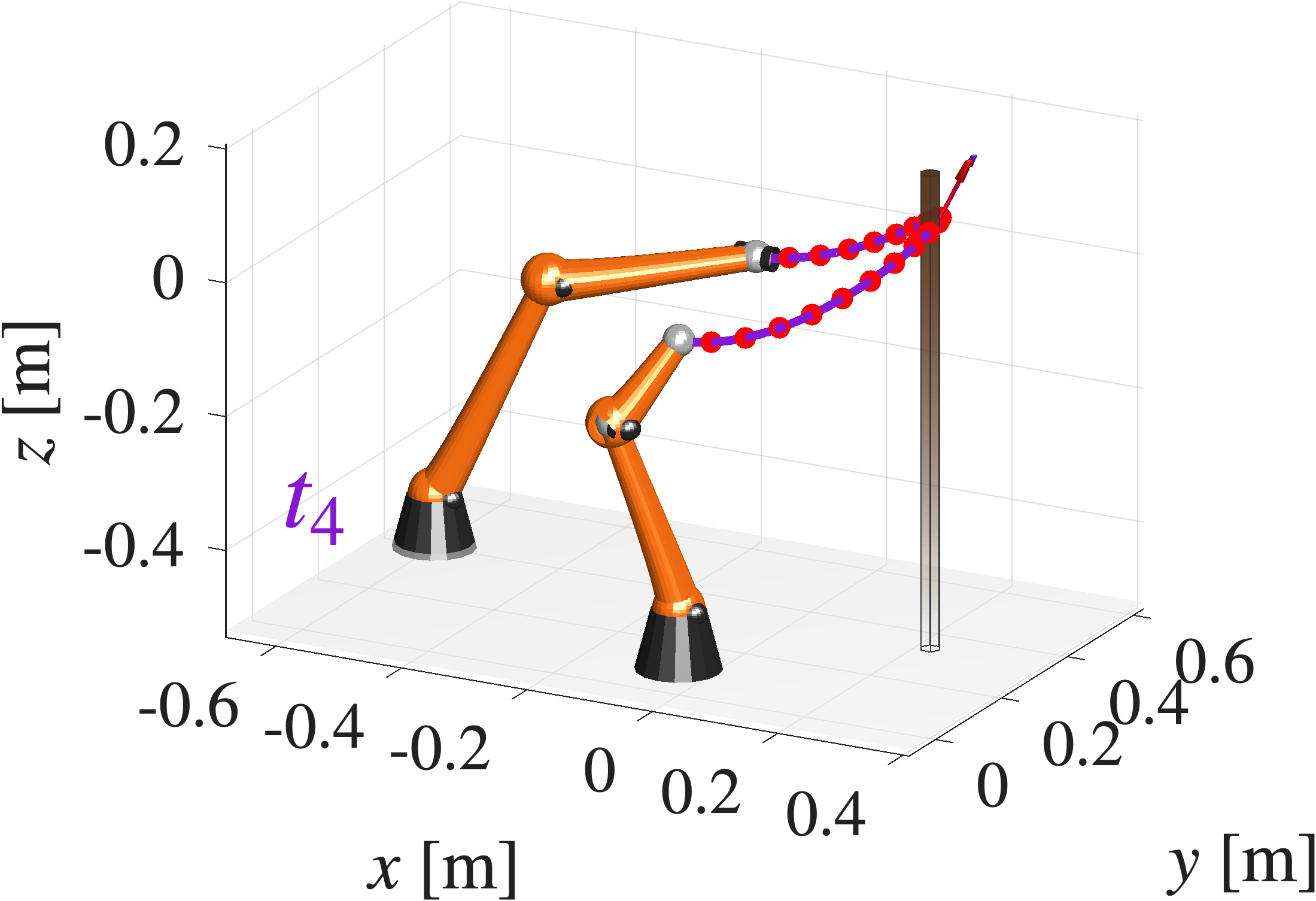}\par\vspace{1mm}
        \includegraphics[width=0.45\linewidth]{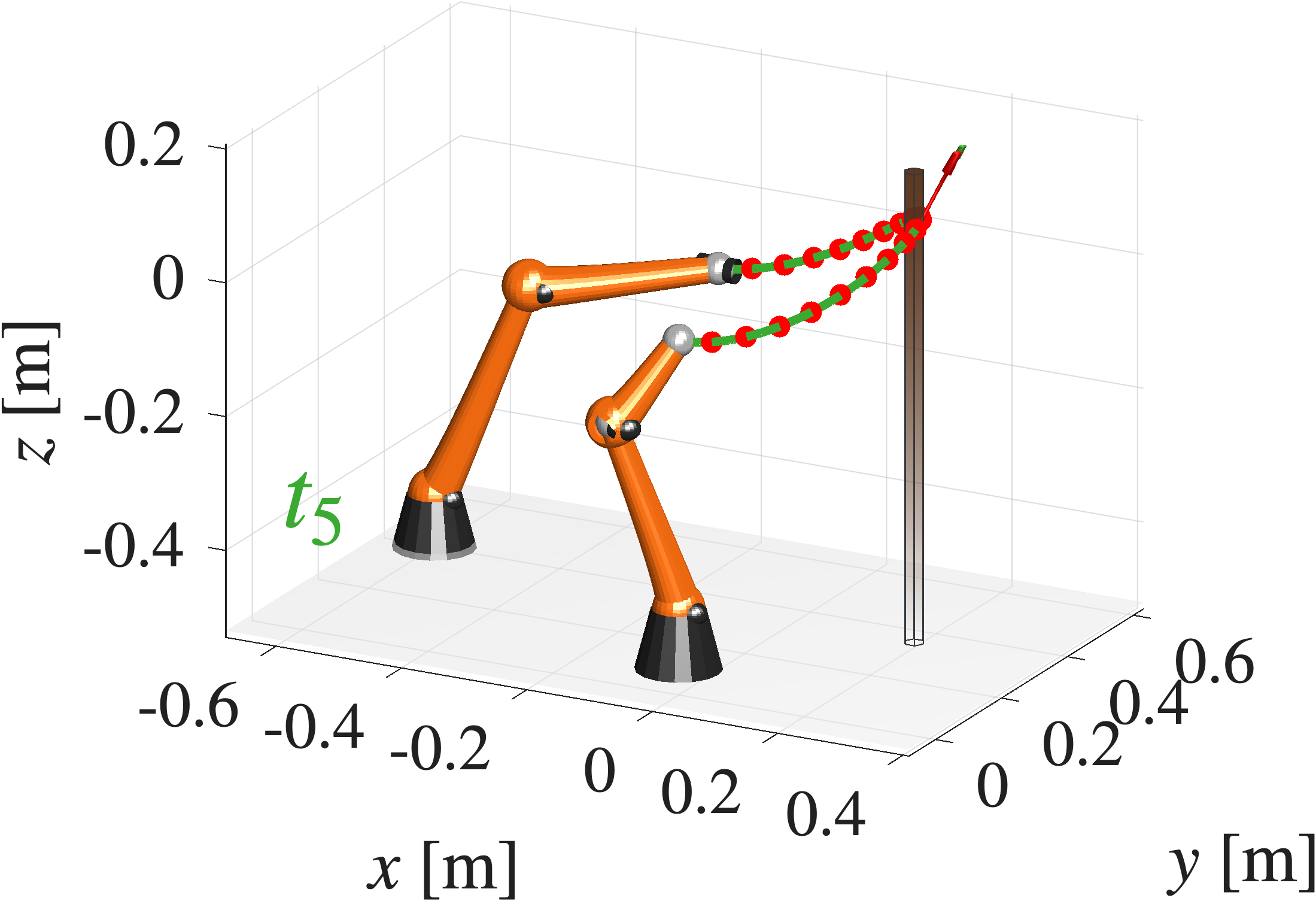}\hfill
        \includegraphics[width=0.45\linewidth]{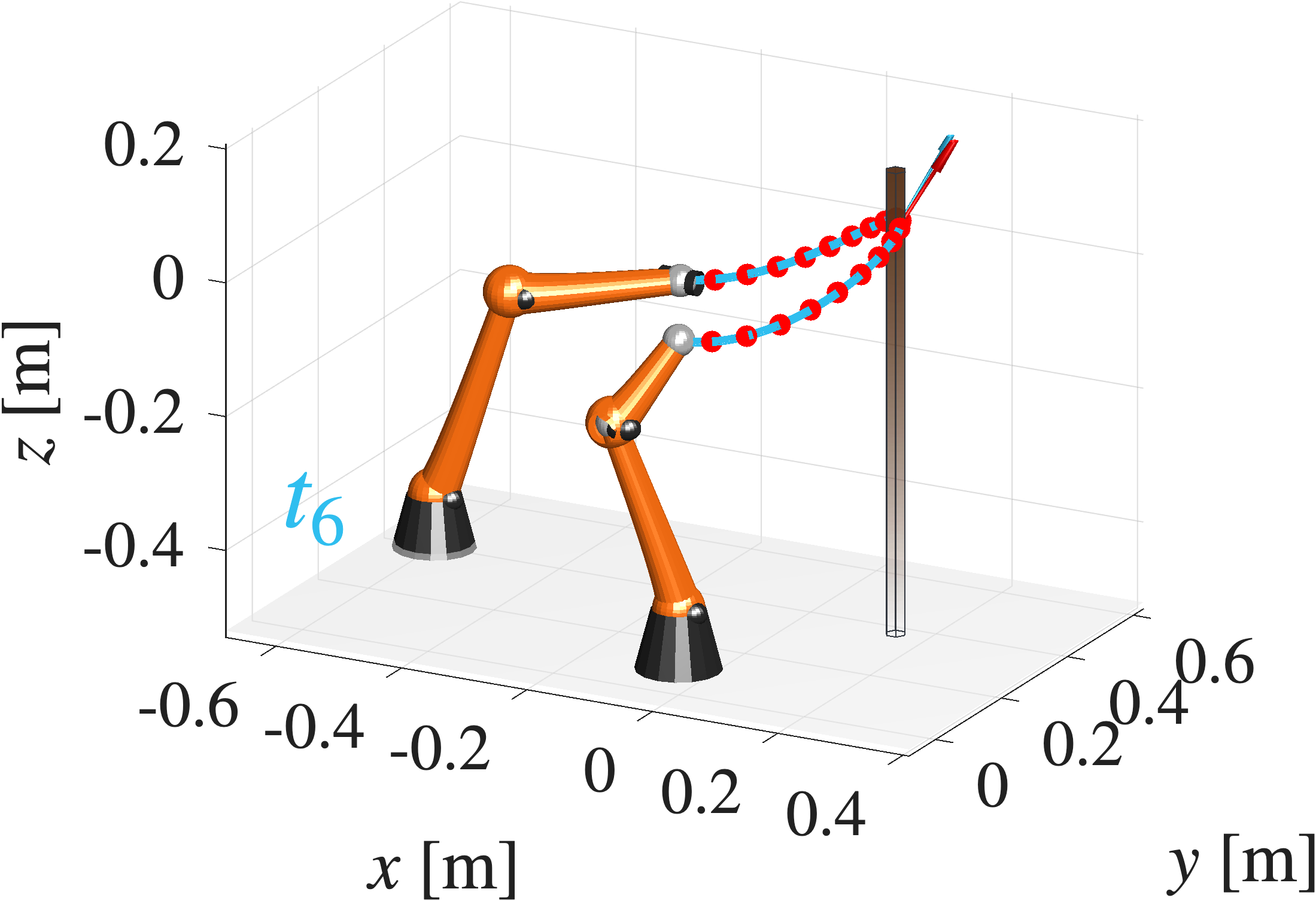}\par\vspace{1mm}
        \includegraphics[width=0.45\linewidth]{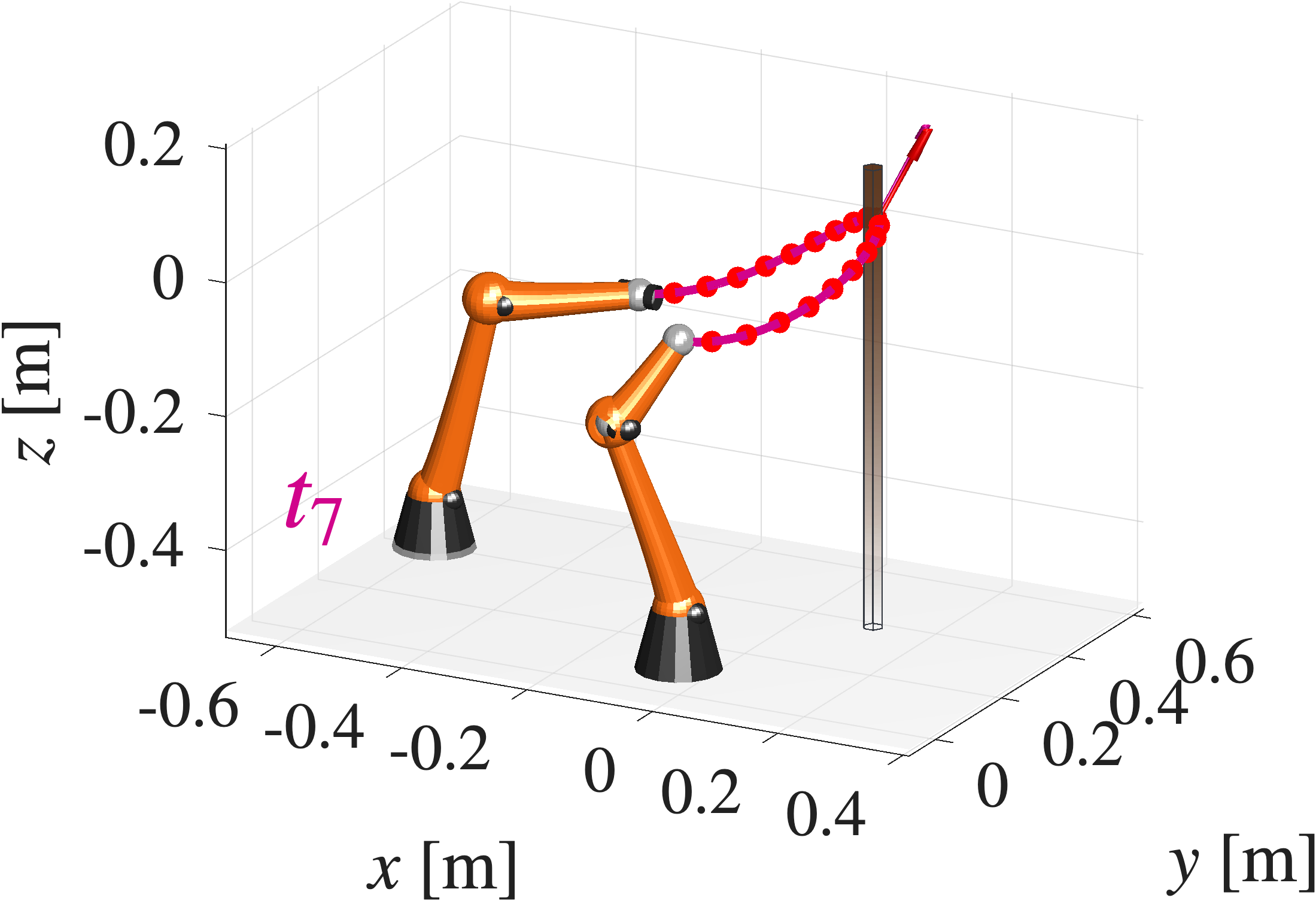}\hfill
        \includegraphics[width=0.45\linewidth]{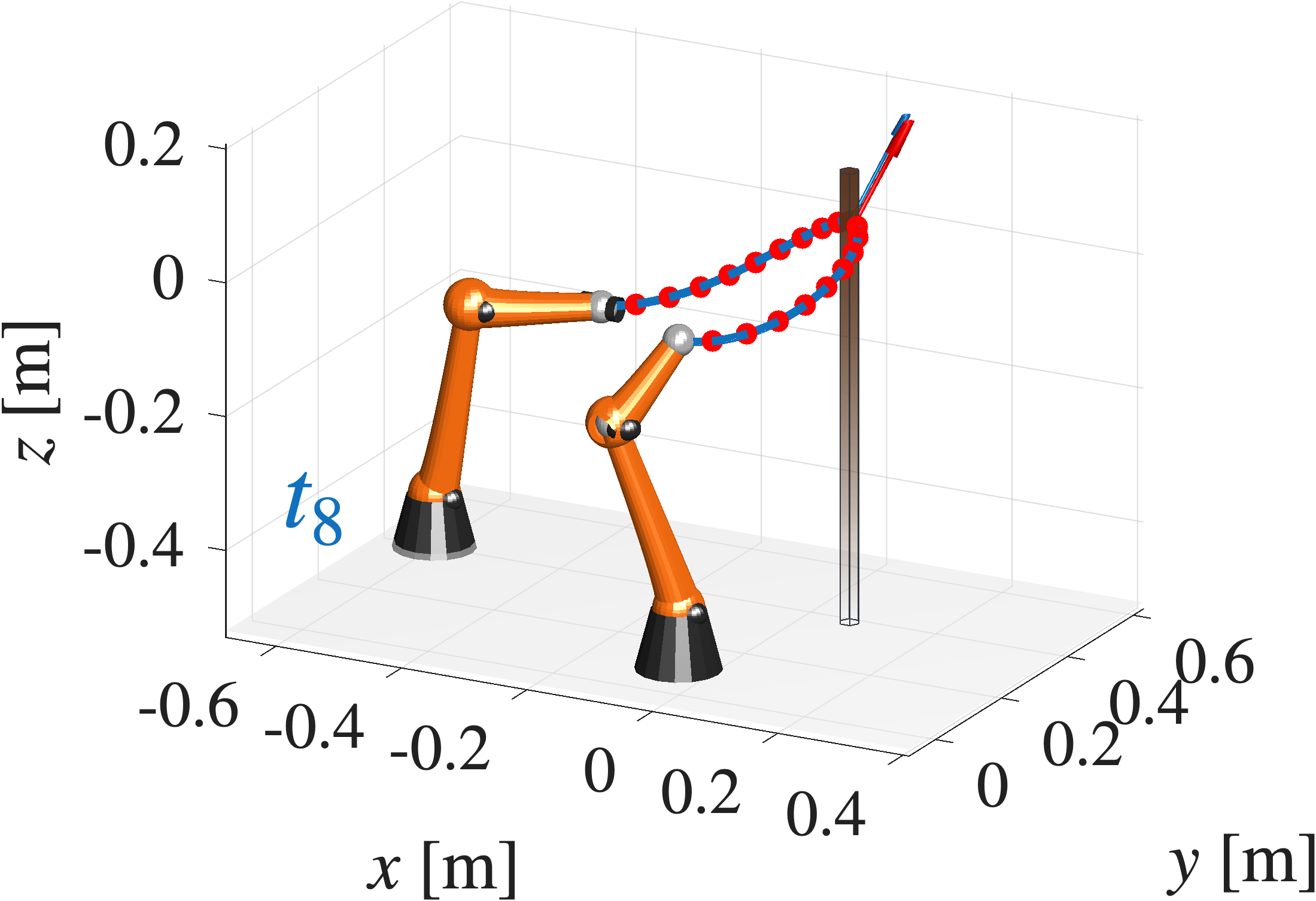}\par\vspace{1mm}
        \includegraphics[width=0.45\linewidth]{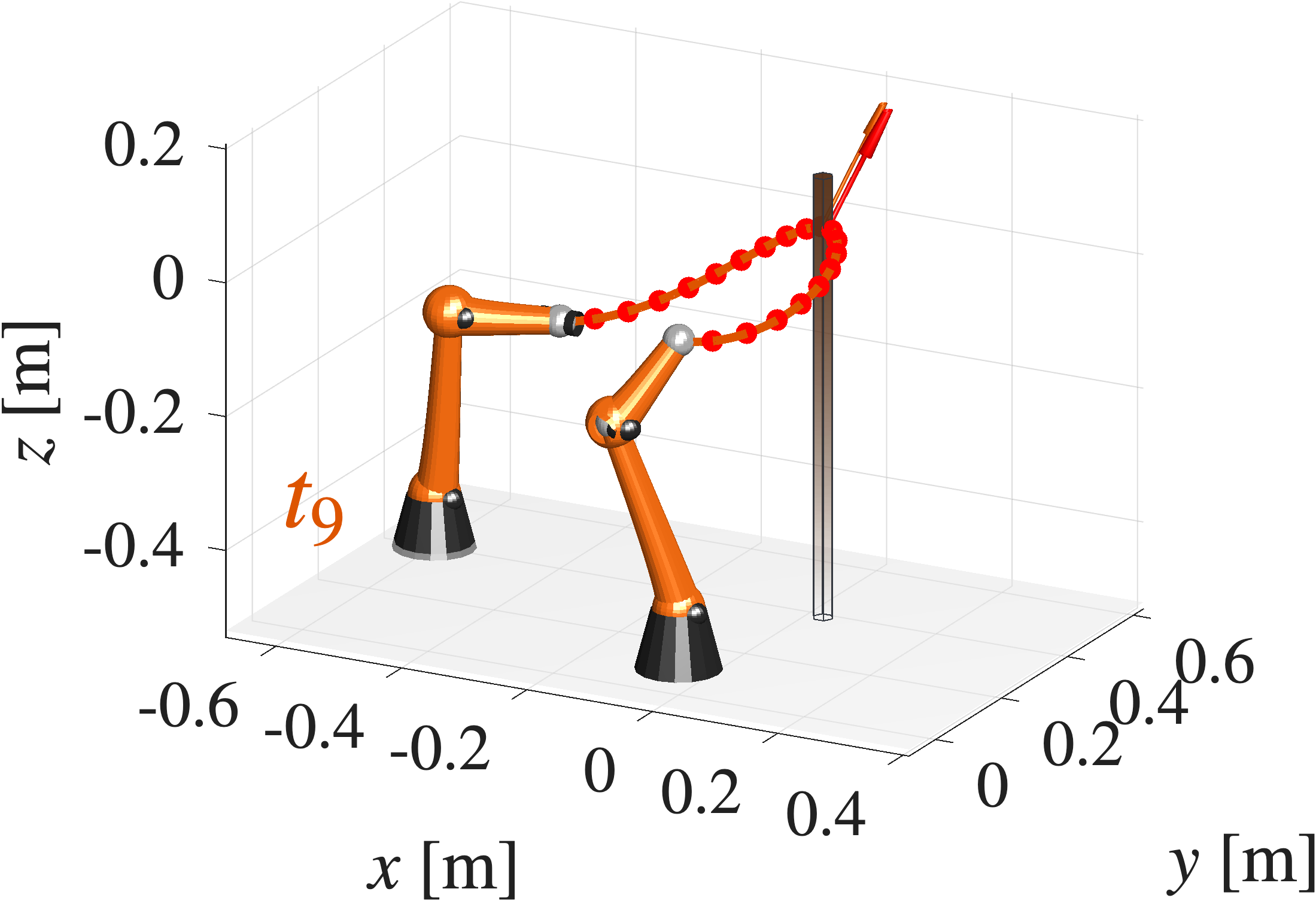}\hfill
        \includegraphics[width=0.45\linewidth]{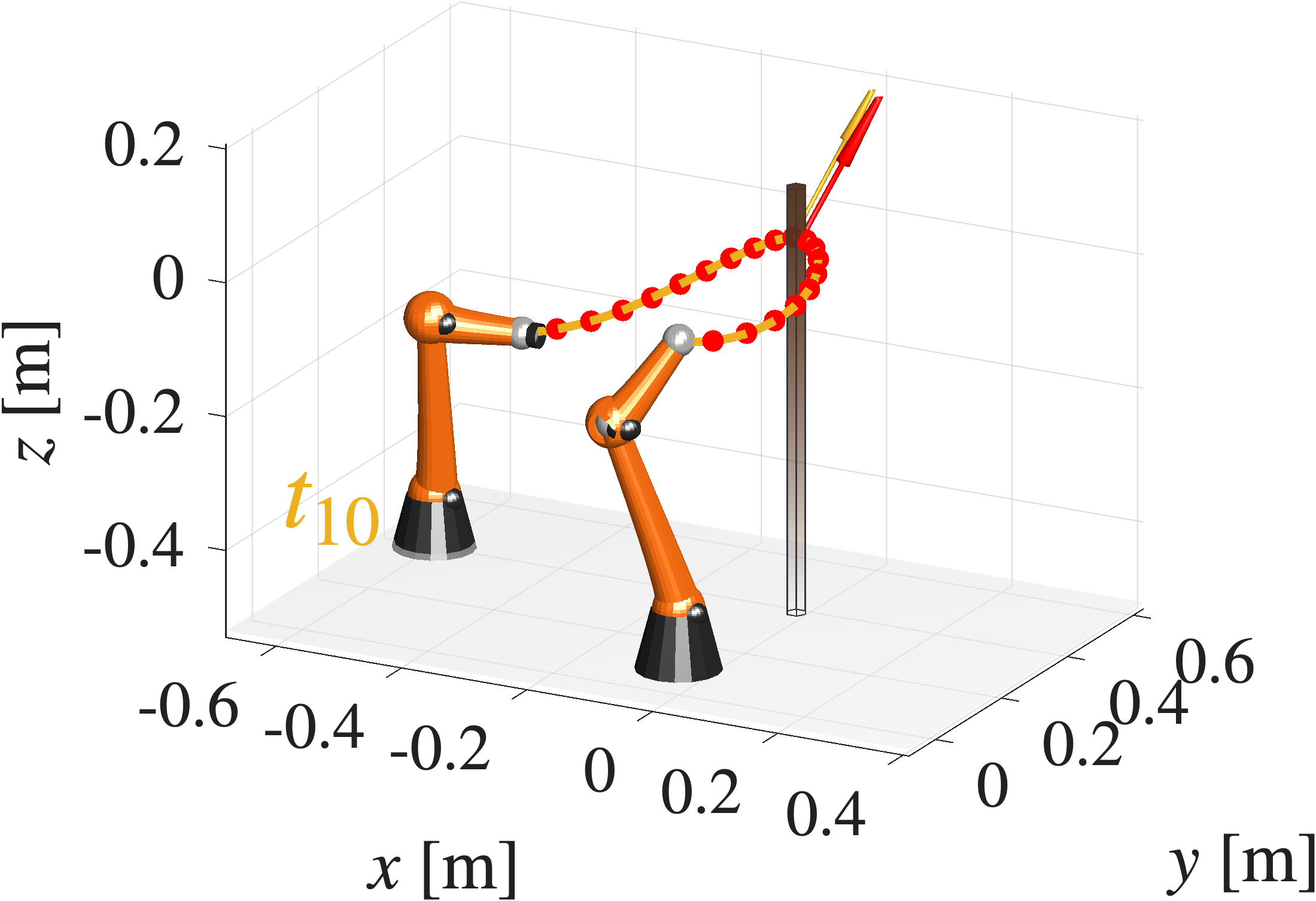}\par
    \end{minipage}%
    \hfill
    \begin{minipage}[t]{0.28\textwidth}
        \centering\offinterlineskip
        \includegraphics[width=0.80\linewidth]{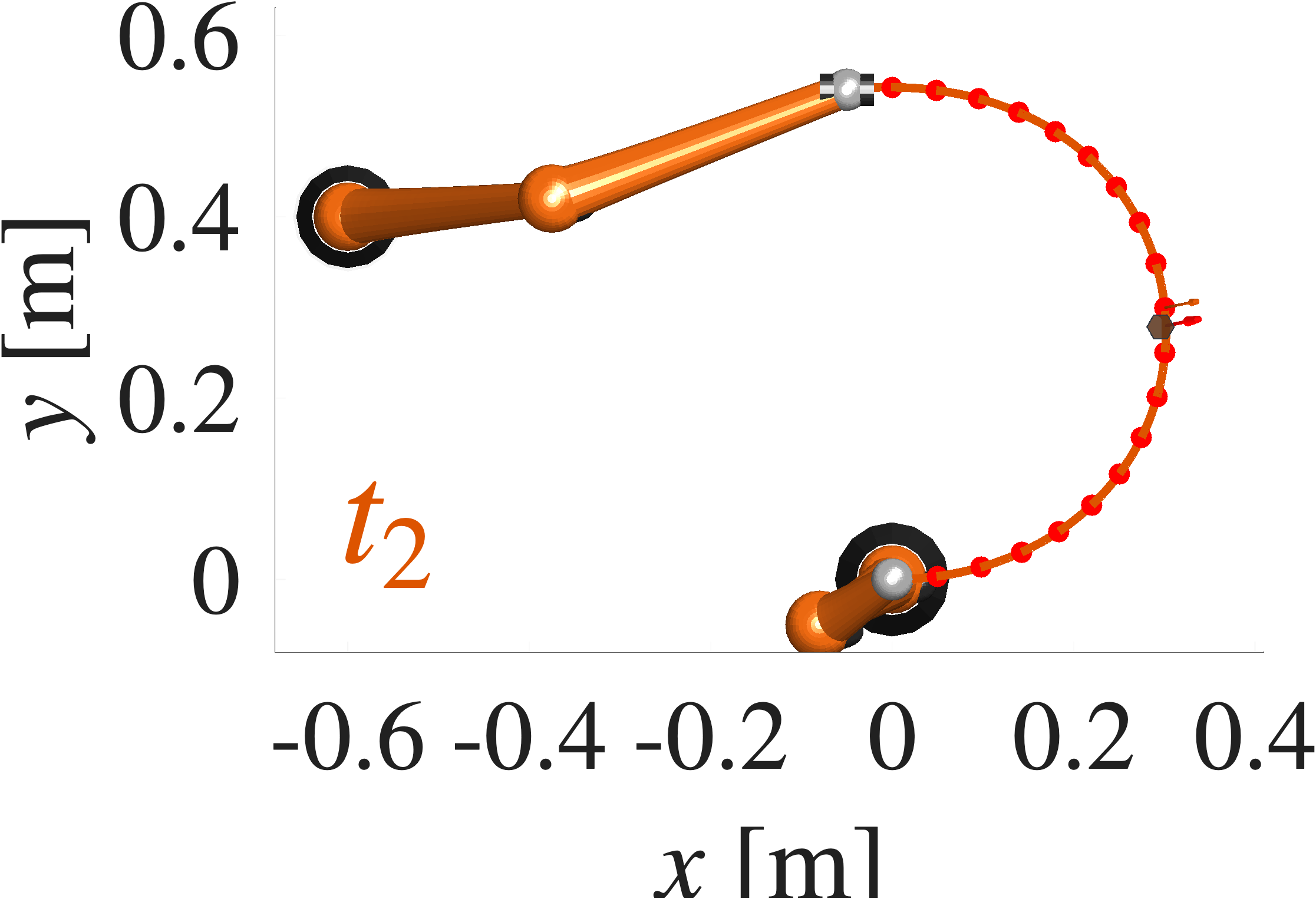}\par\vspace{10mm}
        \includegraphics[width=0.80\linewidth]{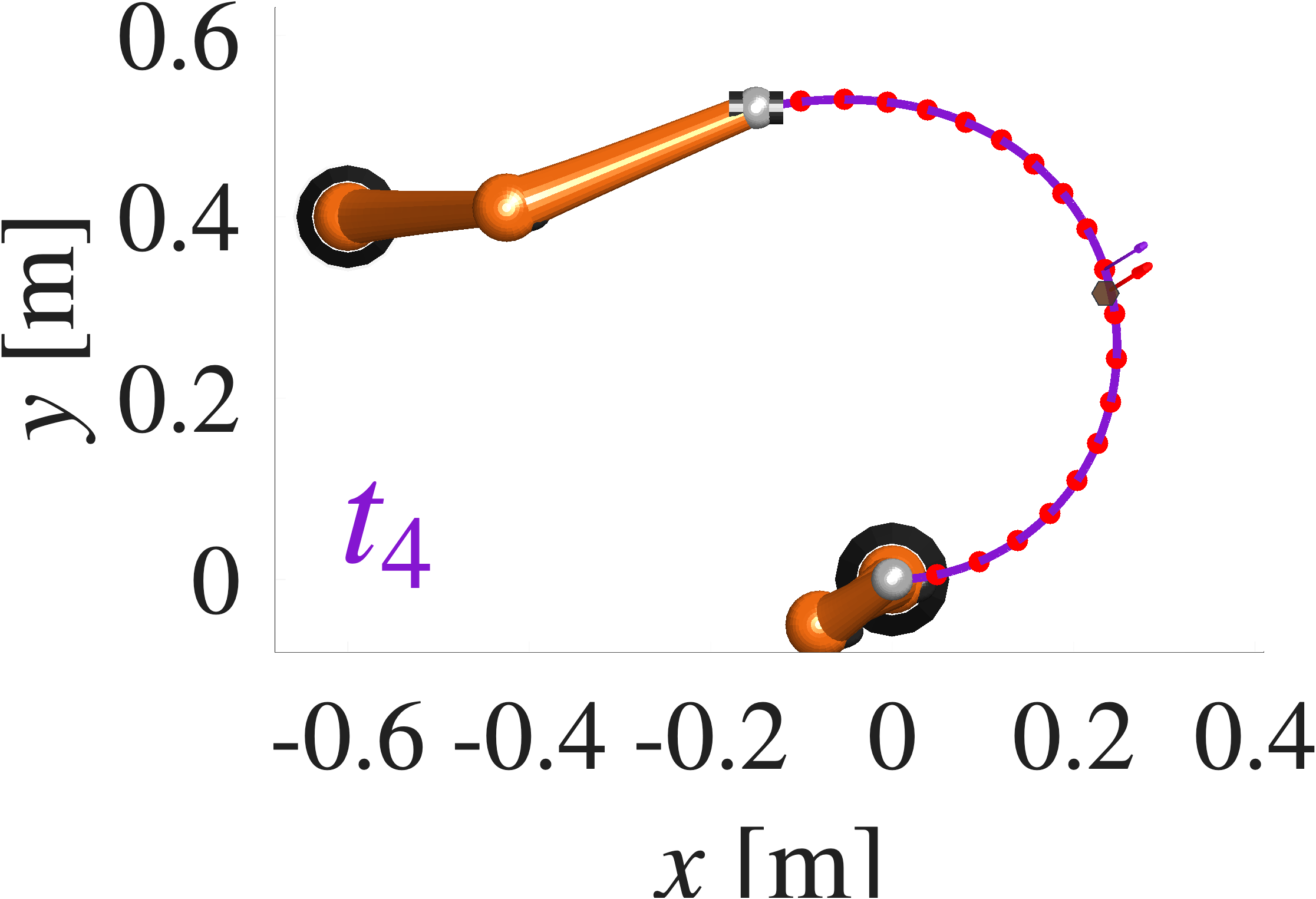}\par\vspace{10mm}
        \includegraphics[width=0.80\linewidth]{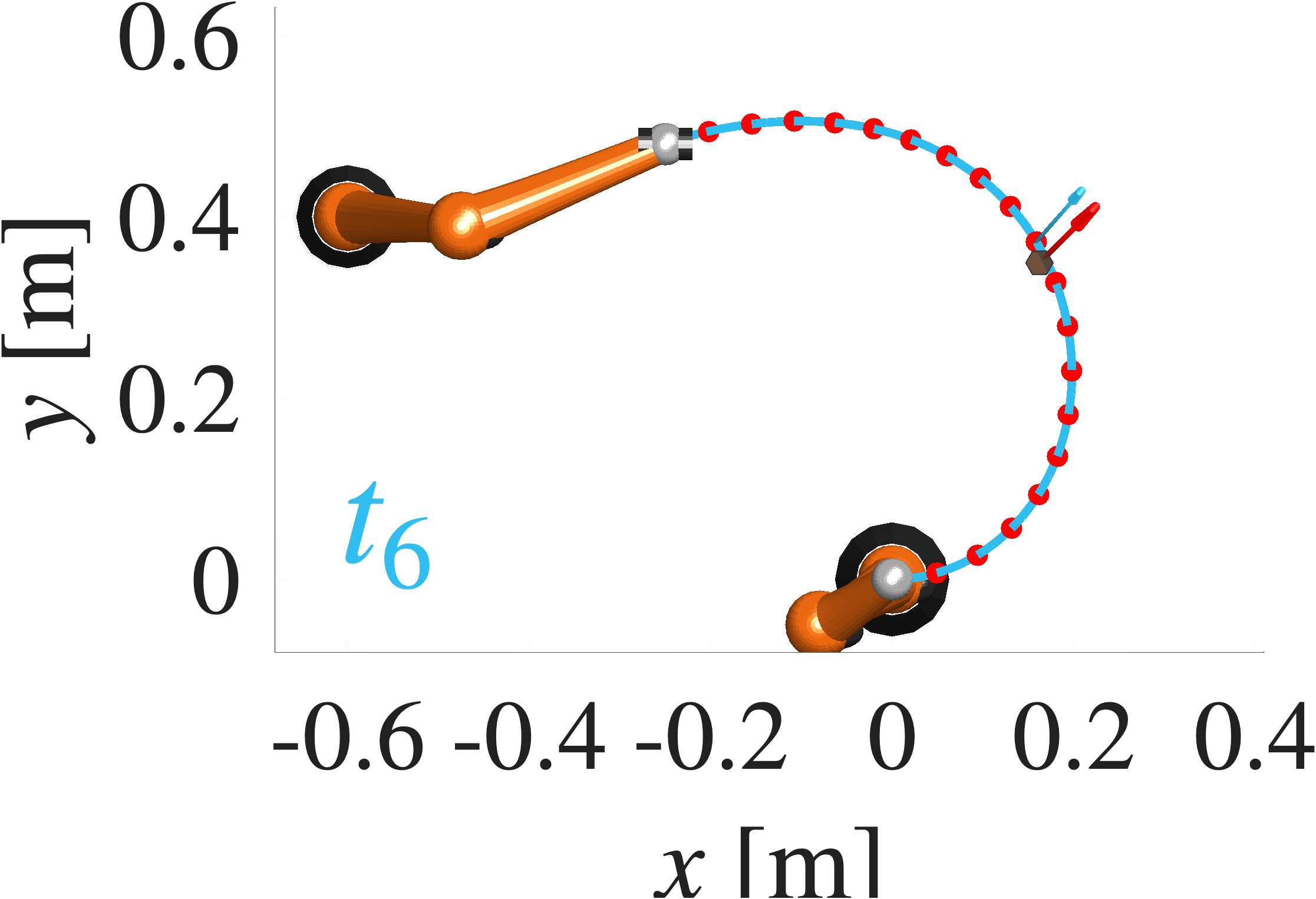}\par\vspace{10mm}
        \includegraphics[width=0.80\linewidth]{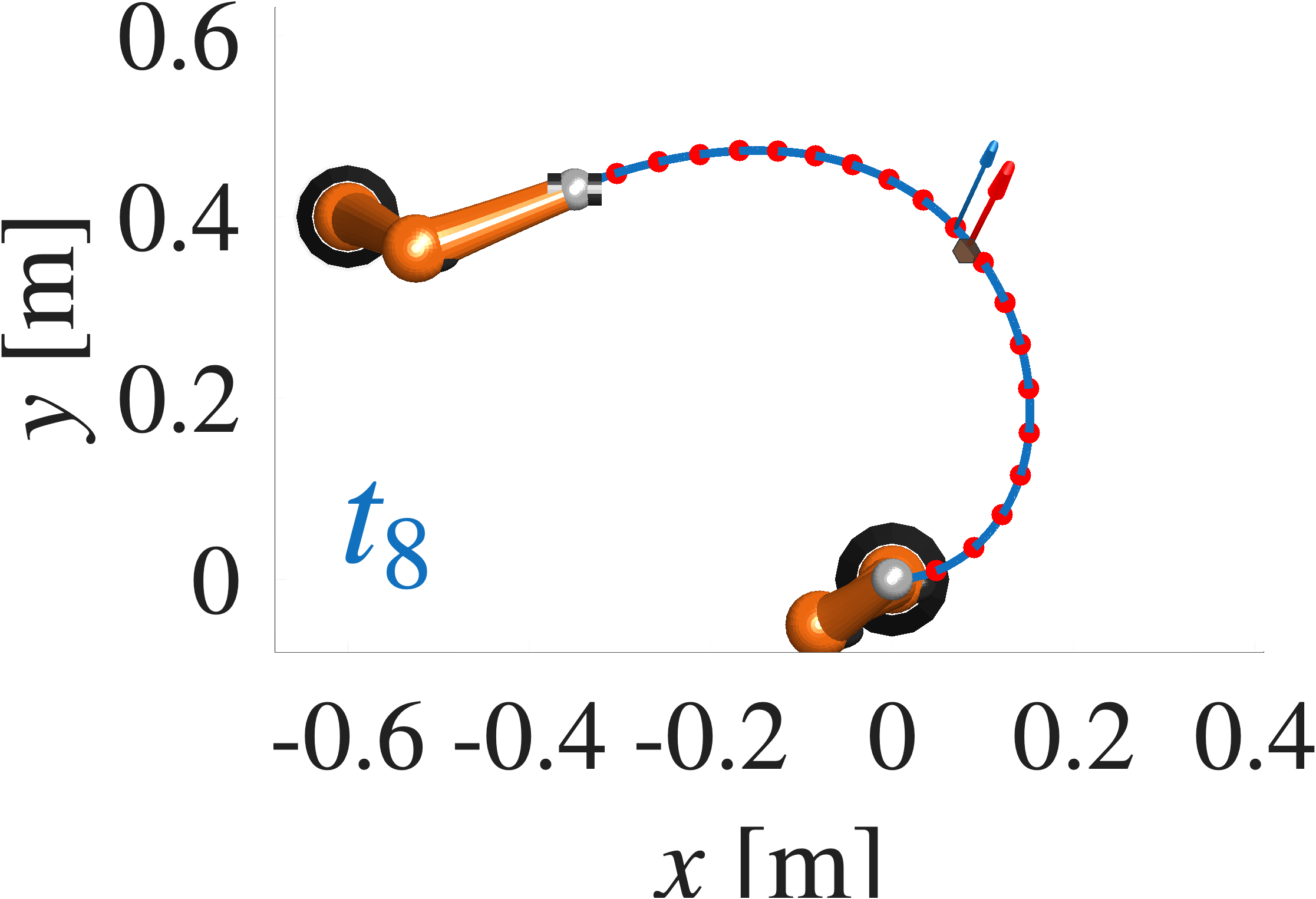}\par\vspace{10mm}
        \includegraphics[width=0.80\linewidth]{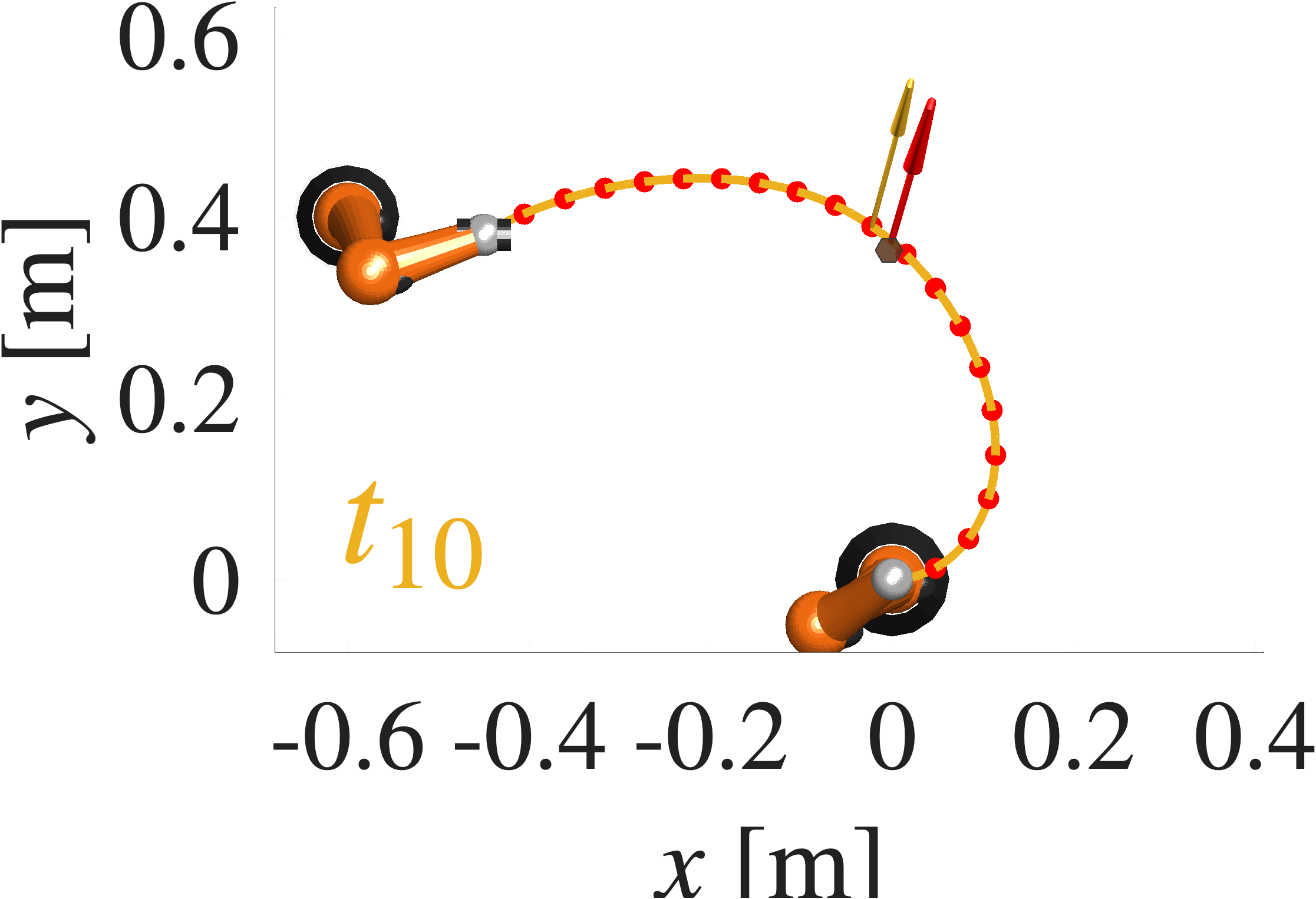}\par
        \vspace{0.8cm}
    \end{minipage}
    
    \includegraphics[width=0.98\linewidth]{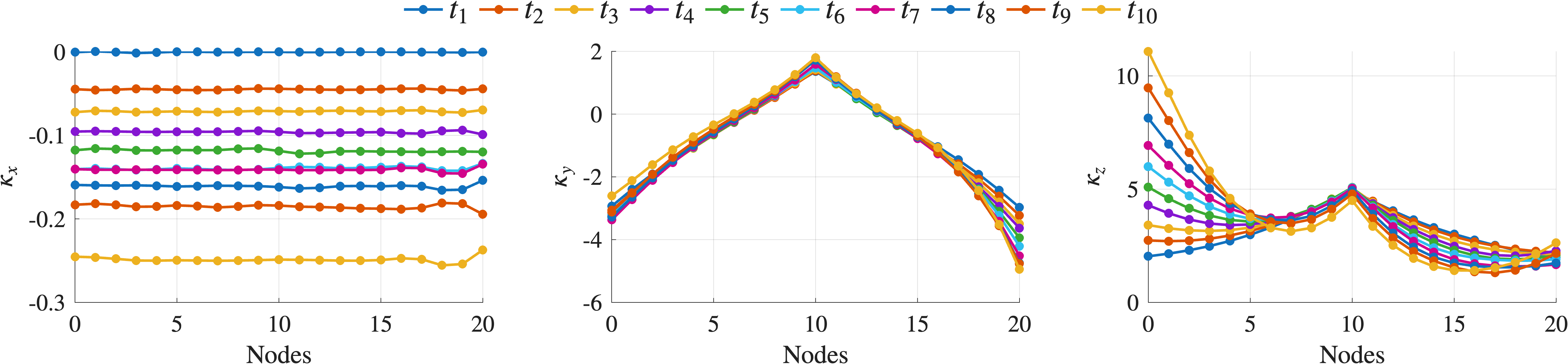}
    \caption{Intermediate configurations of the \textit{pull} trajectory with a force acting at $s=0.5$ and markers measurement noise of $\sigma=1\times10^{-4}$~m. Left: 3D view at steps $t_1$--$t_{10}$. Right: top view at steps $t_2,t_4,t_6,t_8,t_{10}$. Bottom: estimated curvature components $\kappa_x$ (torsion), $\kappa_y, \kappa_z$ (bendings). Through the graph-based algorithm, node poses at discrete points along the rod and the external force are estimated and shown in red at each step.}
    \label{fig:pull-traj-1}
\end{figure*}

 \begin{figure*}[p]
    \centering
    \begin{minipage}[t]{0.68\textwidth}
        \centering\offinterlineskip
        \includegraphics[width=0.45\linewidth]{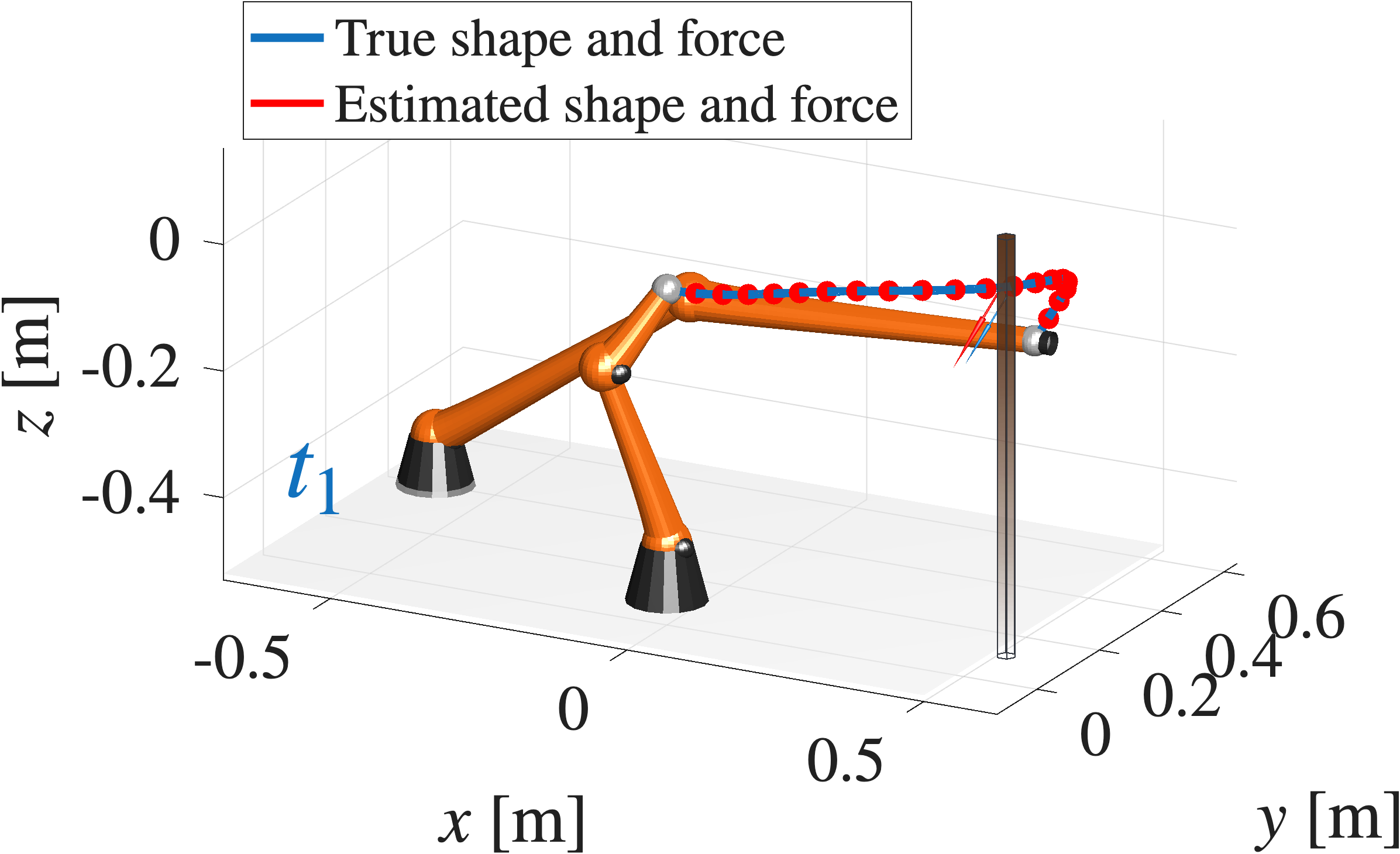}\hfill
        \includegraphics[width=0.45\linewidth]{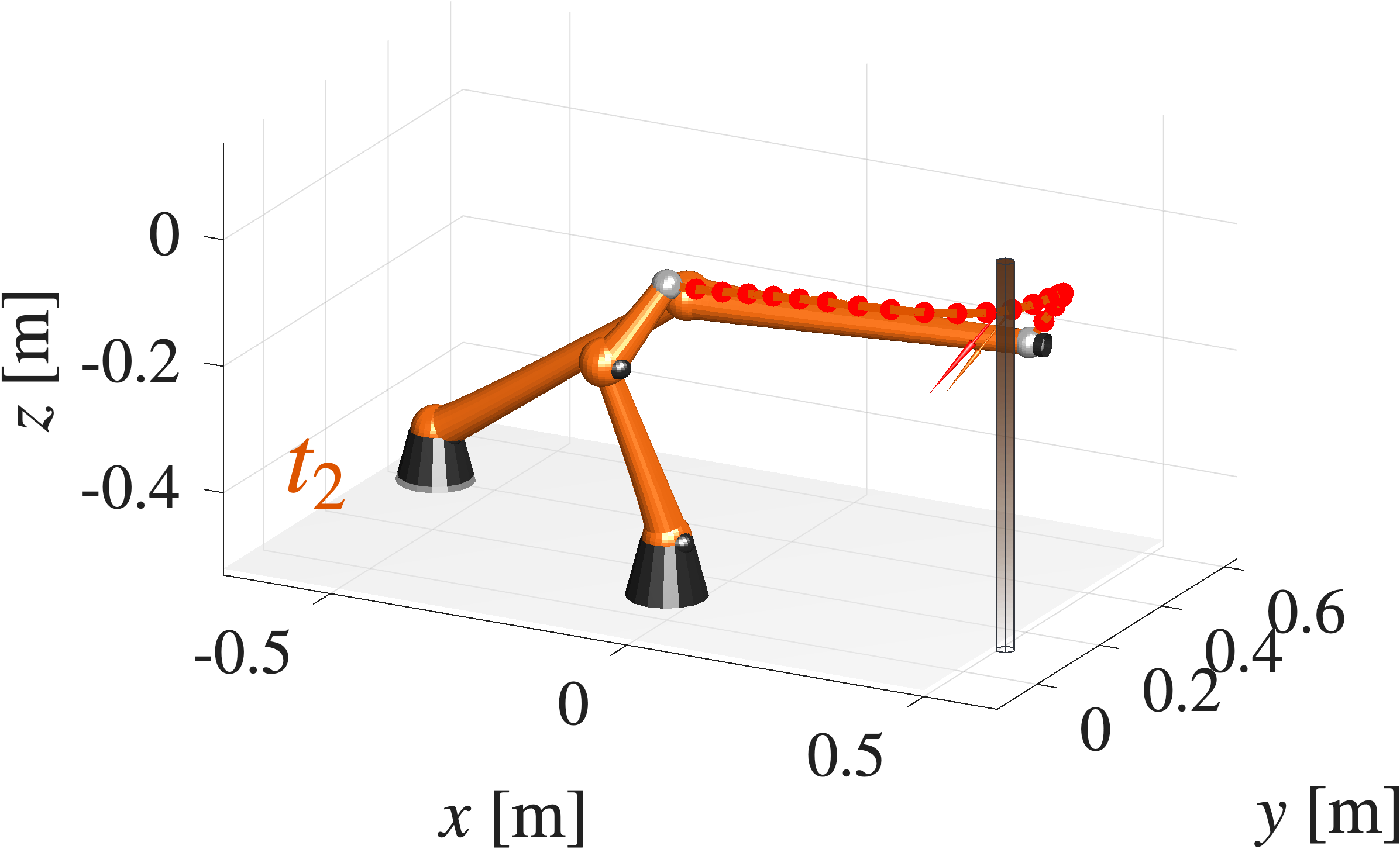}\par\vspace{2mm}
        \includegraphics[width=0.45\linewidth]{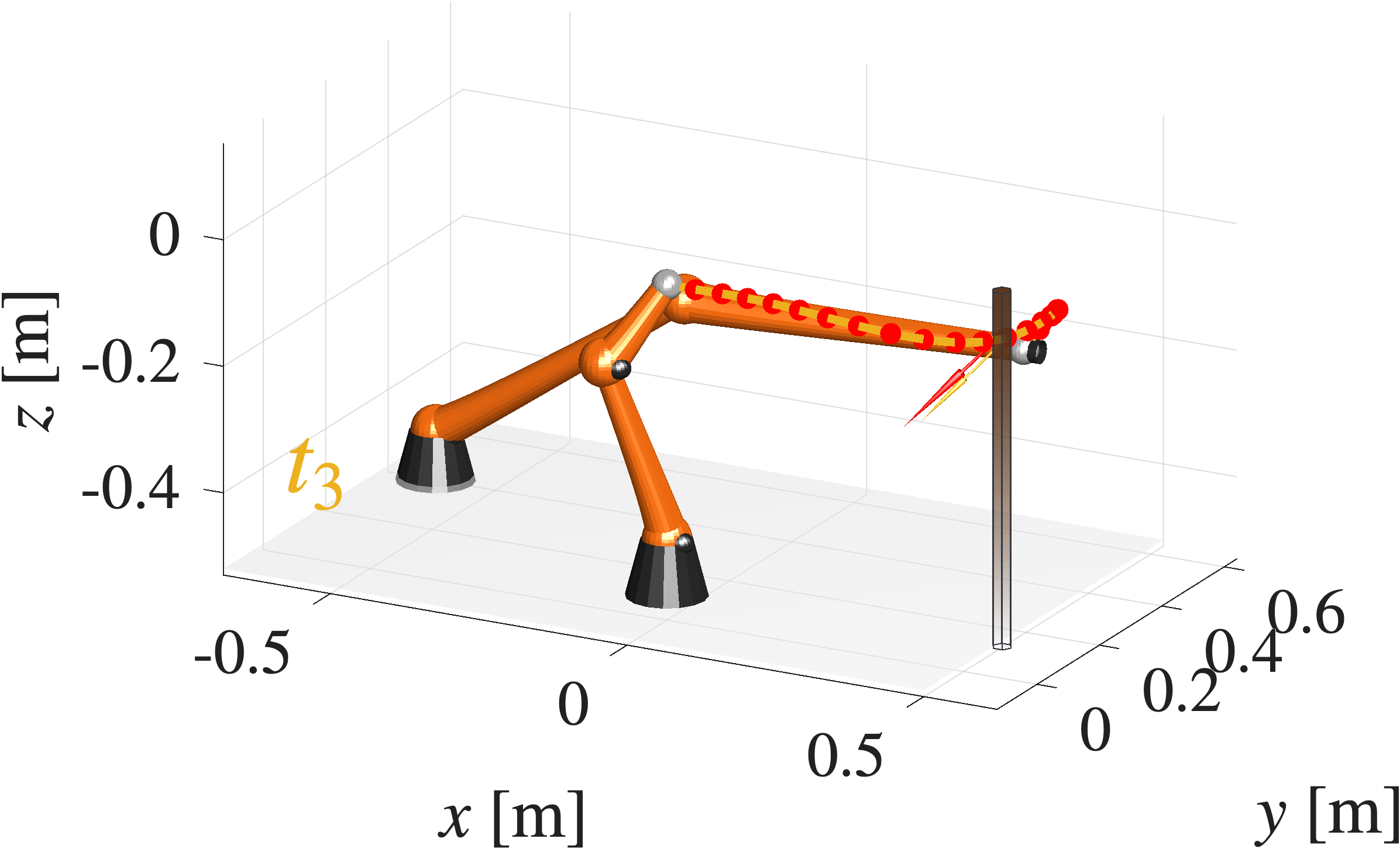}\hfill
        \includegraphics[width=0.45\linewidth]{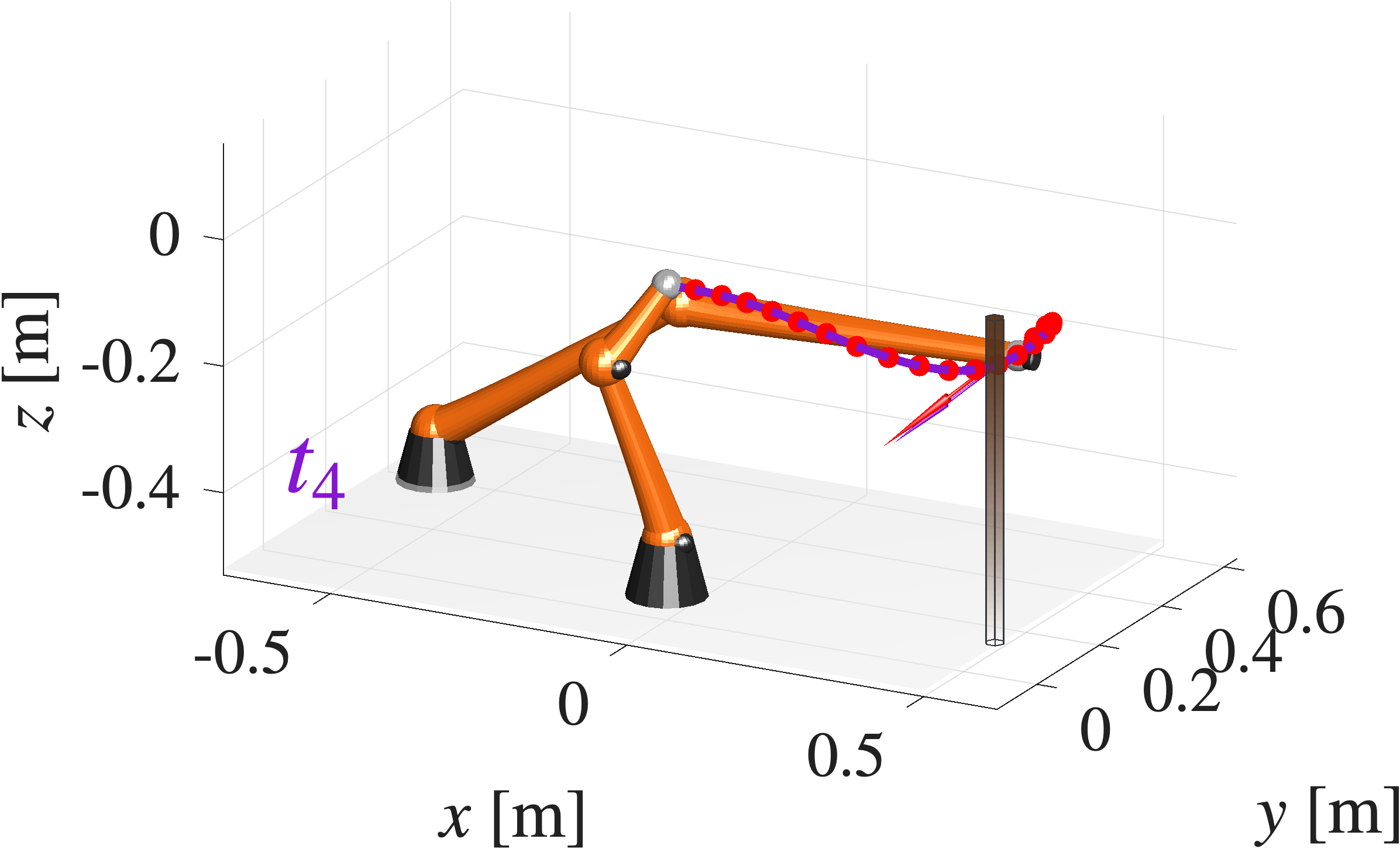}\par\vspace{2mm}
        \includegraphics[width=0.45\linewidth]{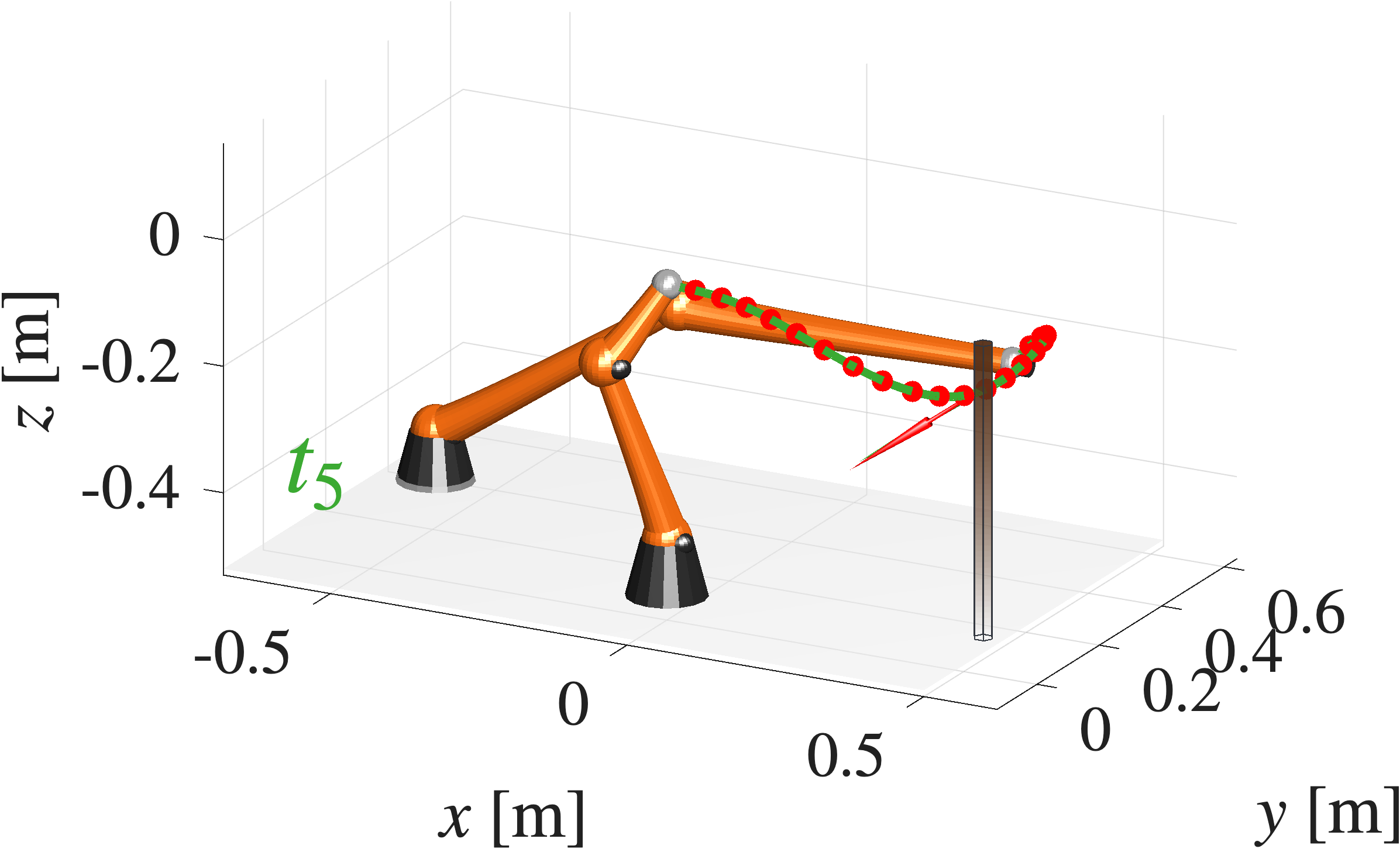}\hfill
        \includegraphics[width=0.45\linewidth]{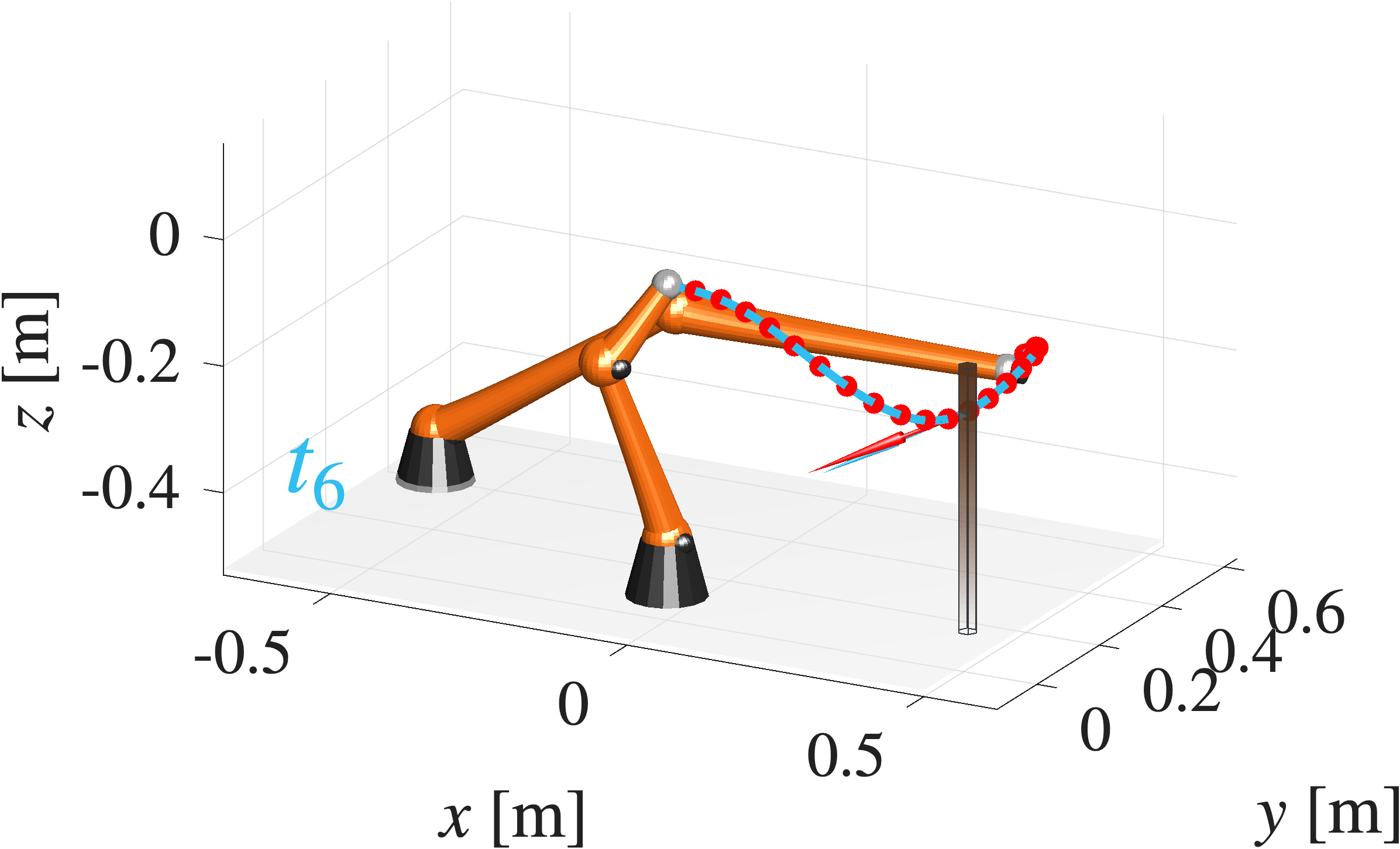}\par\vspace{2mm}
        \includegraphics[width=0.45\linewidth]{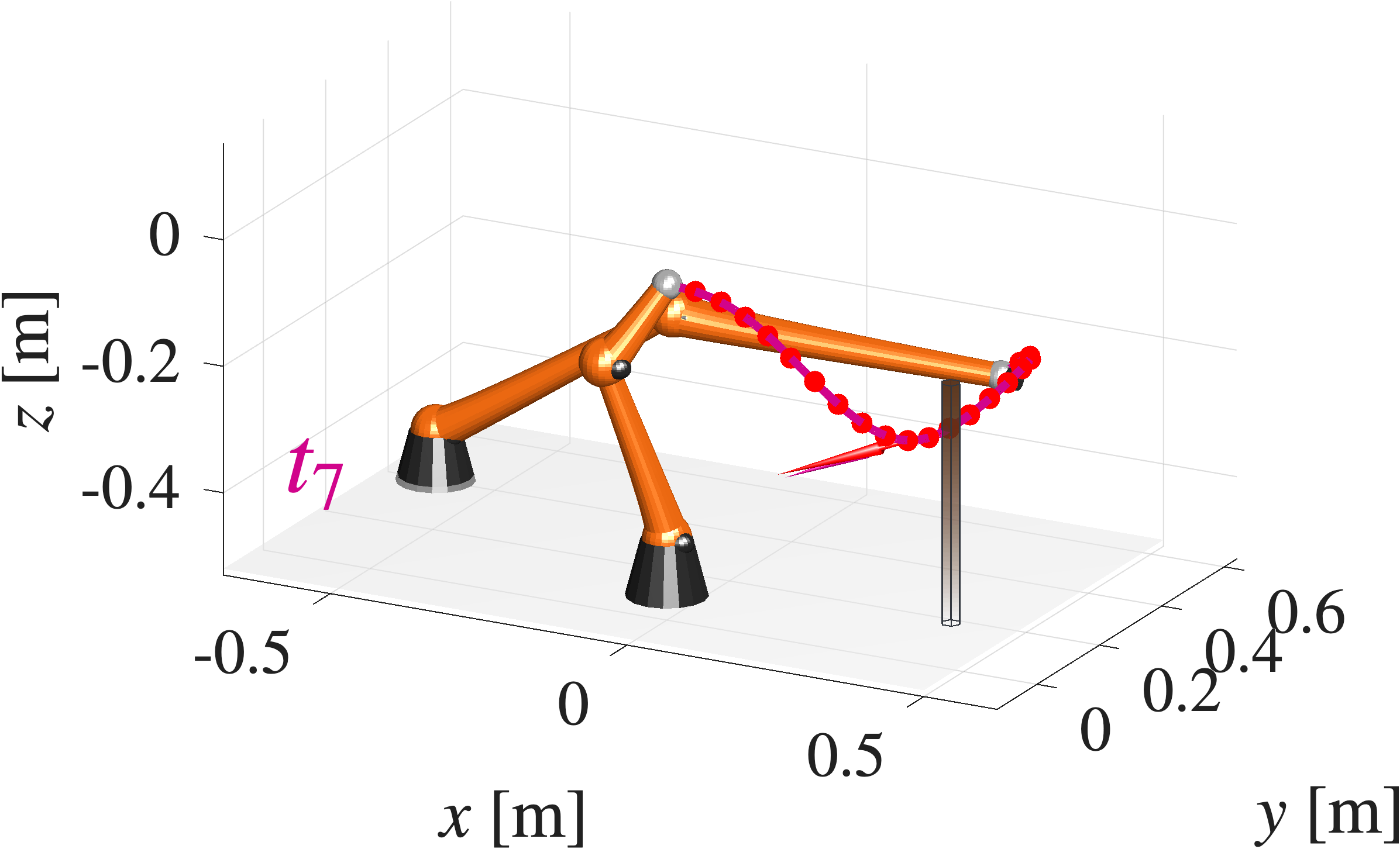}\hfill
        \includegraphics[width=0.45\linewidth]{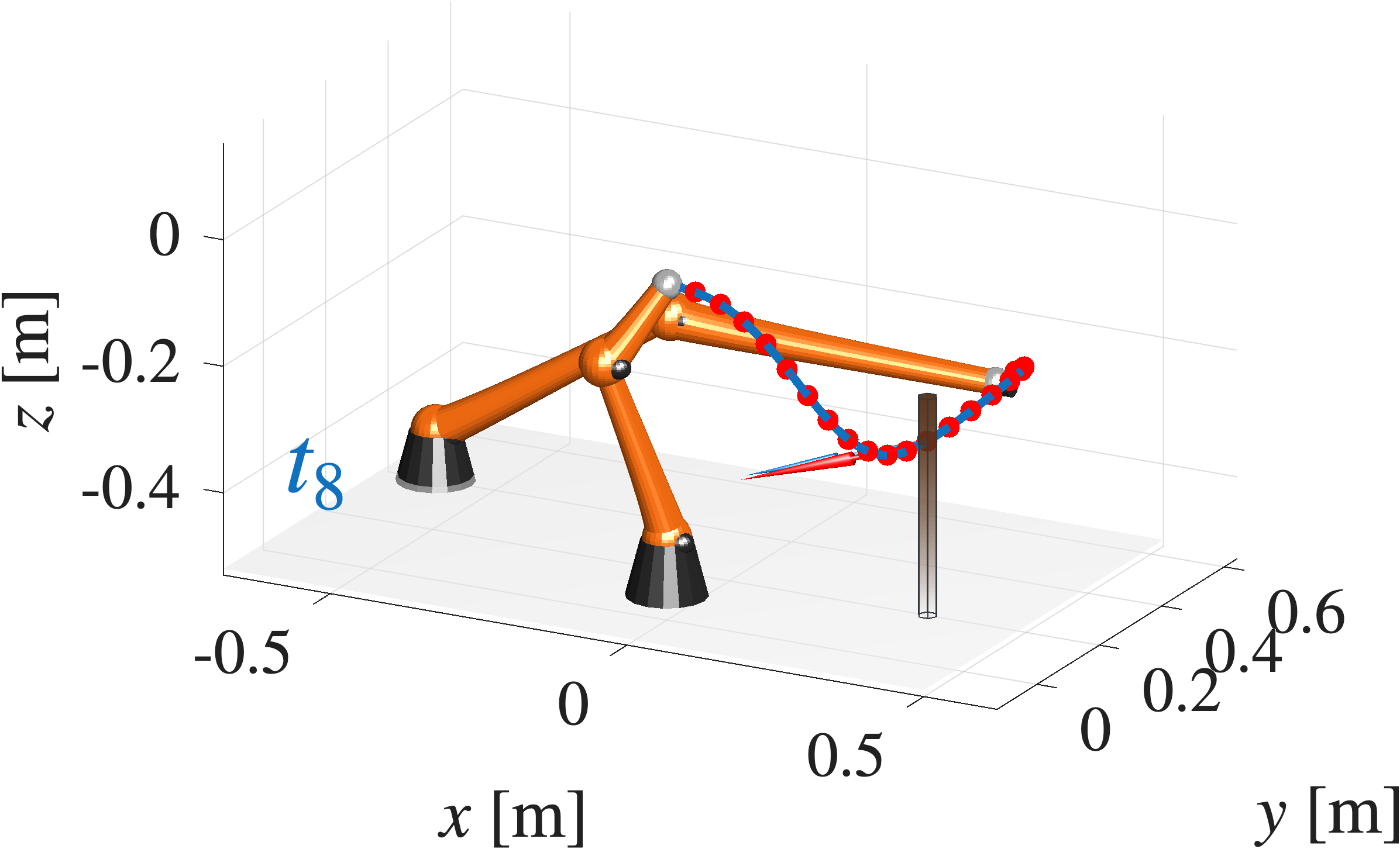}\par\vspace{2mm} 
        \includegraphics[width=0.45\linewidth]{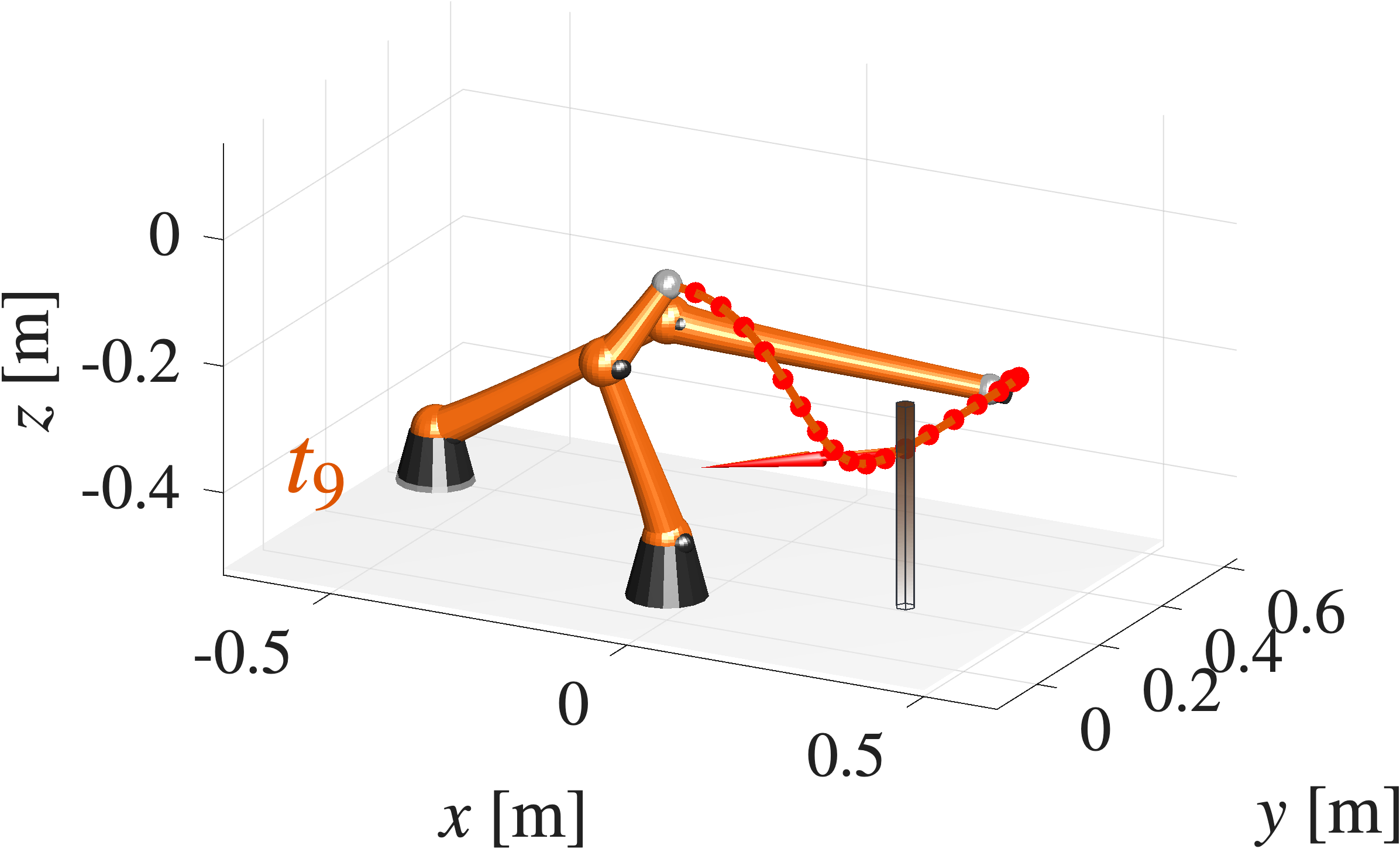}\hfill
        \includegraphics[width=0.45\linewidth]{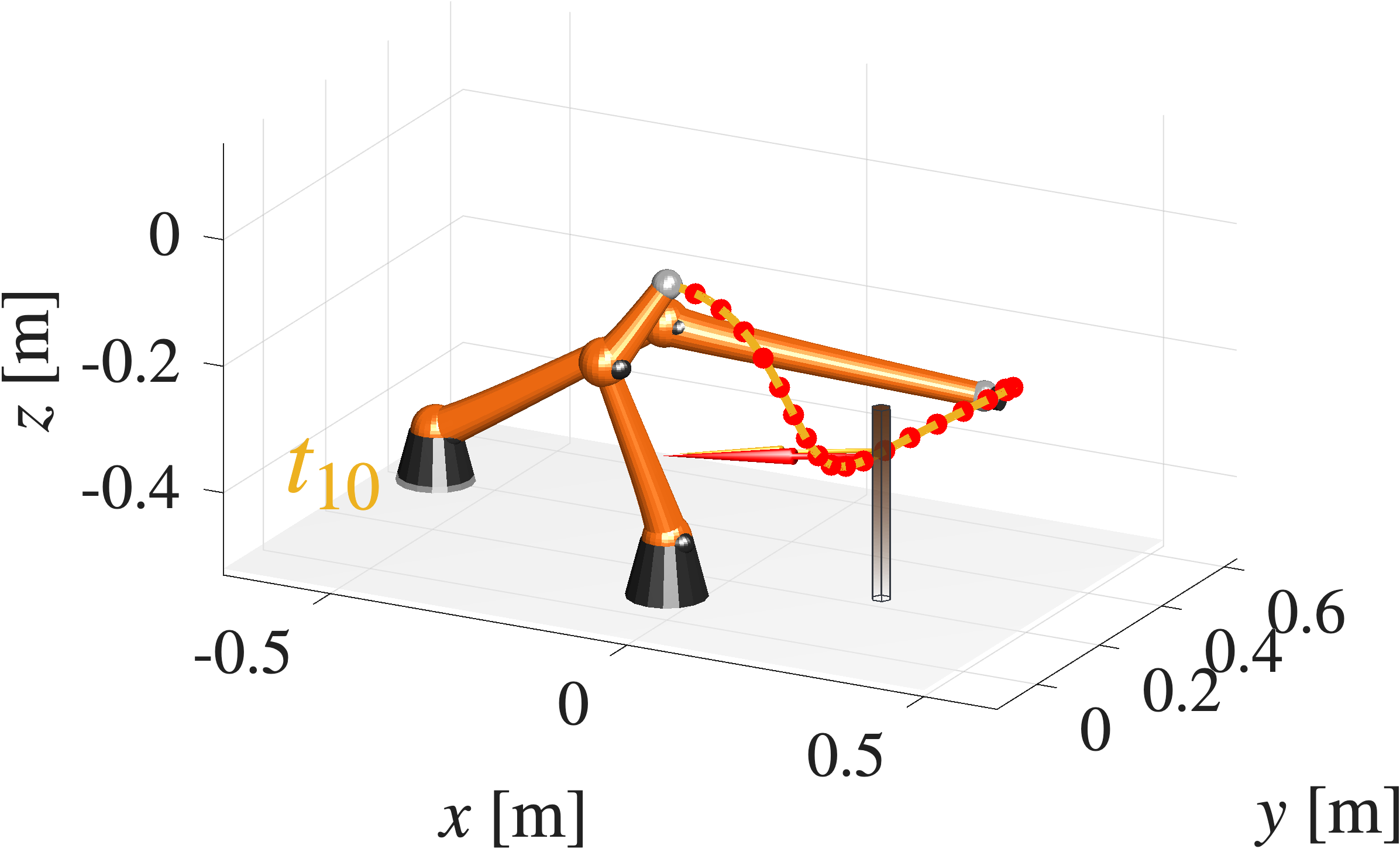}\par
    \end{minipage}%
    \hfill
    \begin{minipage}[t]{0.28\textwidth}
        \centering\offinterlineskip
        \includegraphics[width=0.80\linewidth]{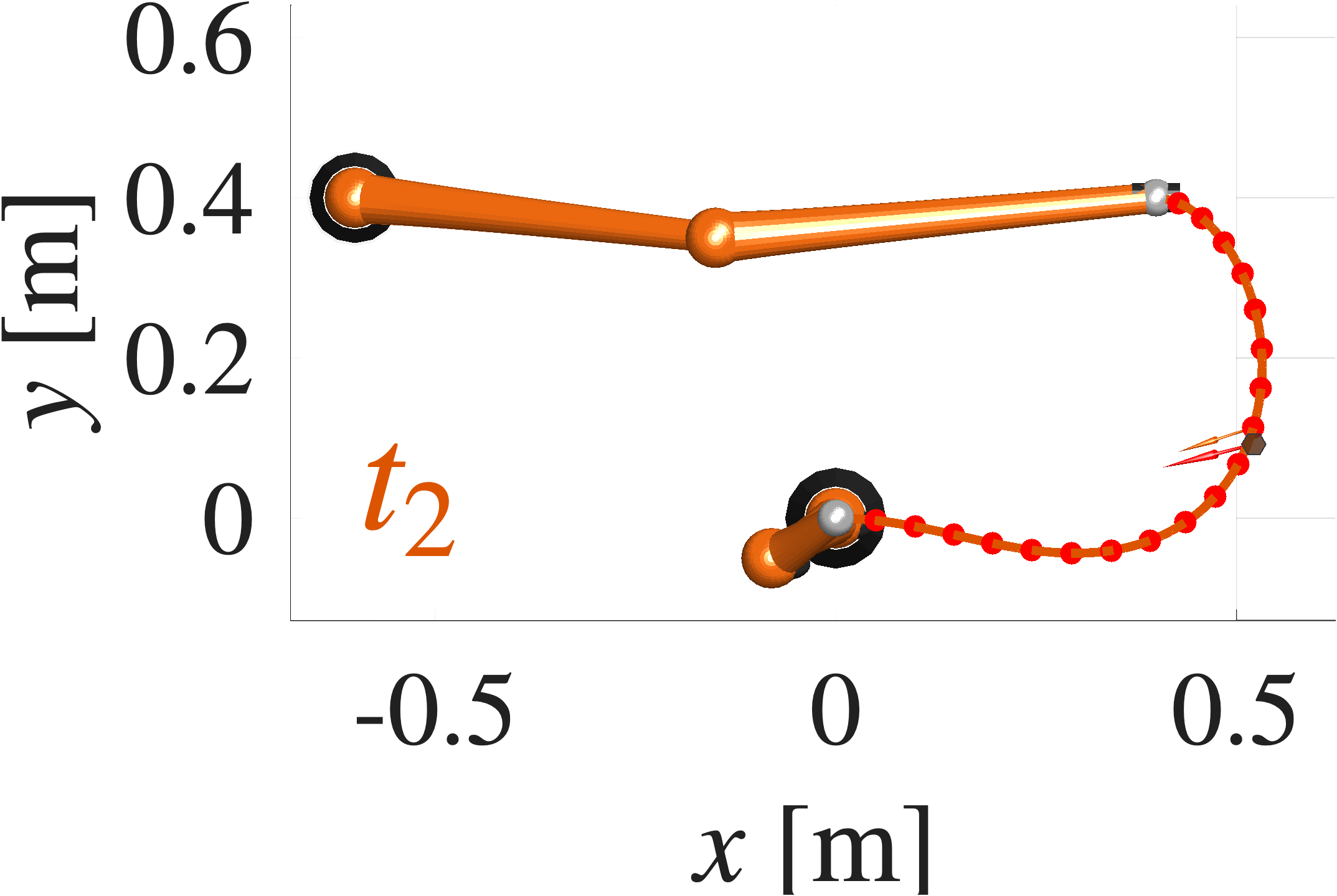}\par\vspace{10mm}
        \includegraphics[width=0.80\linewidth]{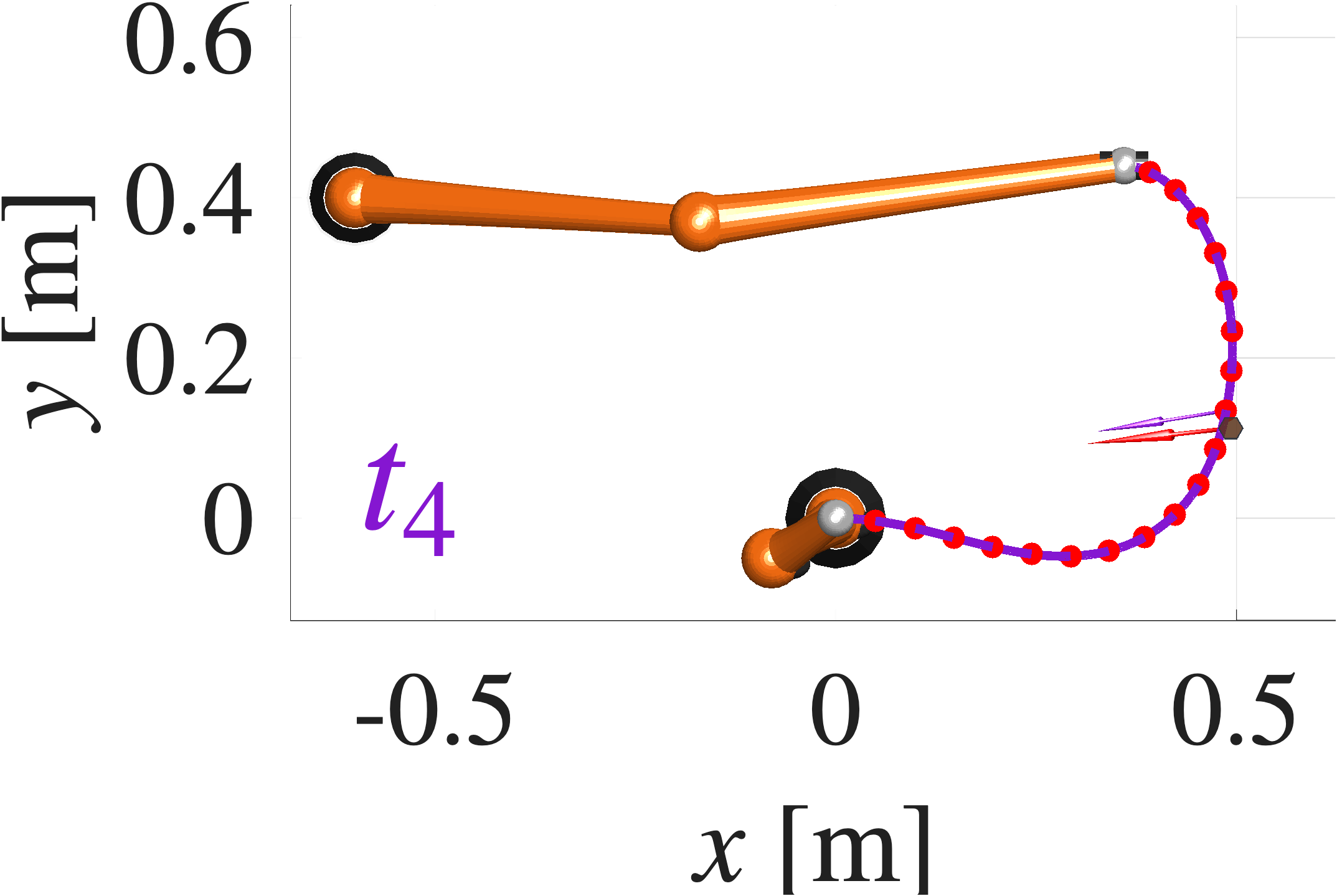}\par\vspace{10mm}
        \includegraphics[width=0.80\linewidth]{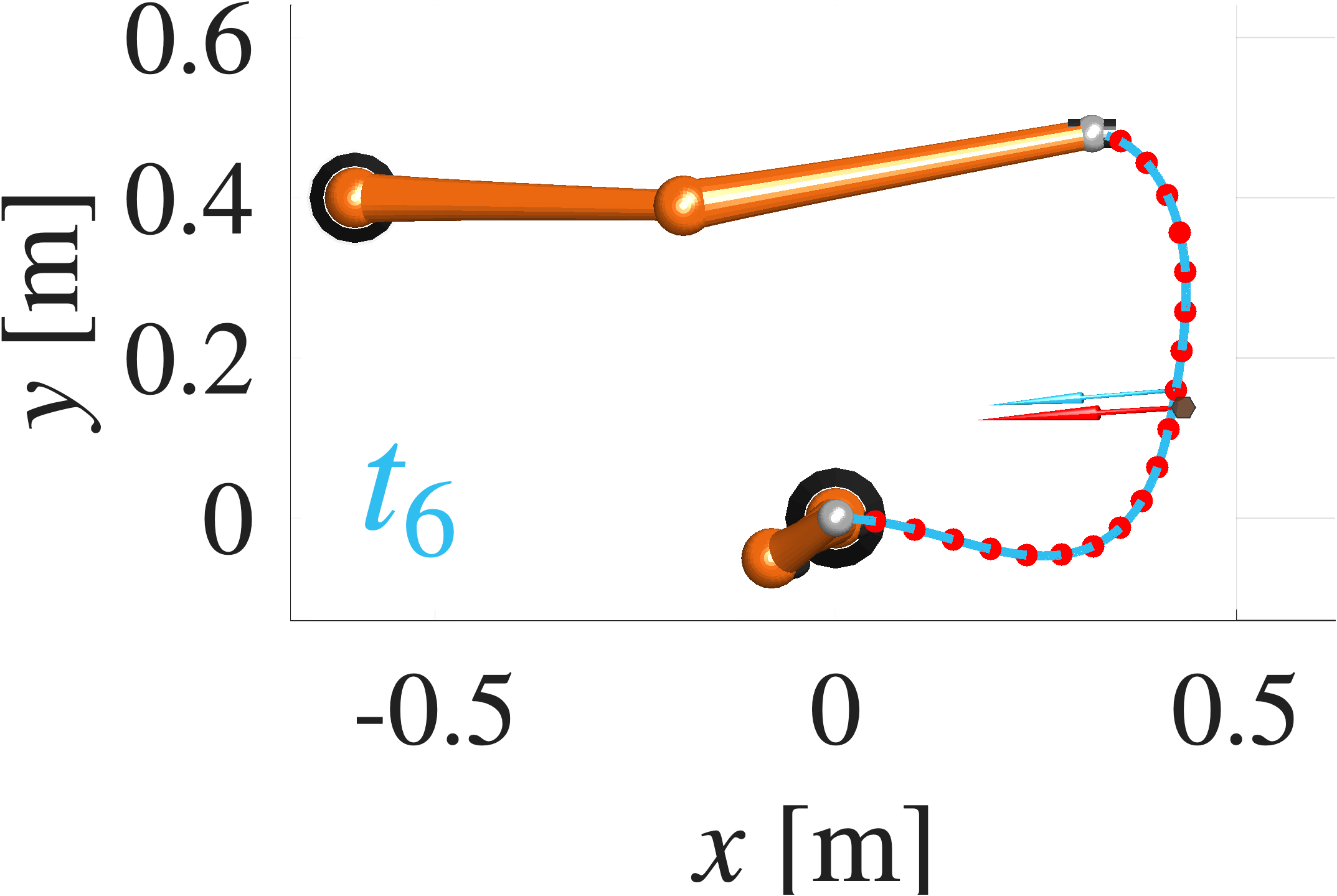}\par\vspace{10mm}
        \includegraphics[width=0.80\linewidth]{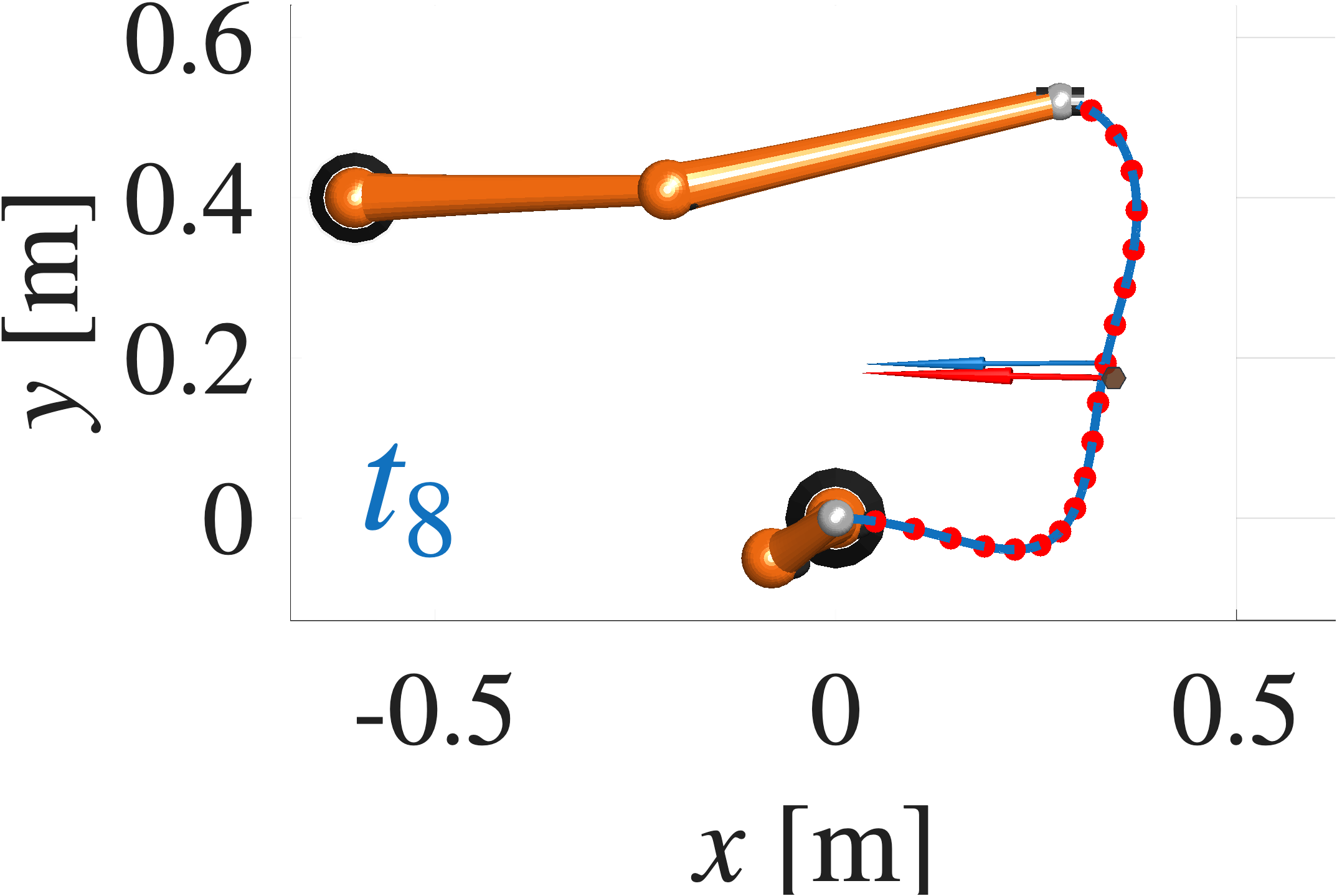}\par\vspace{10mm}
        \includegraphics[width=0.80\linewidth]{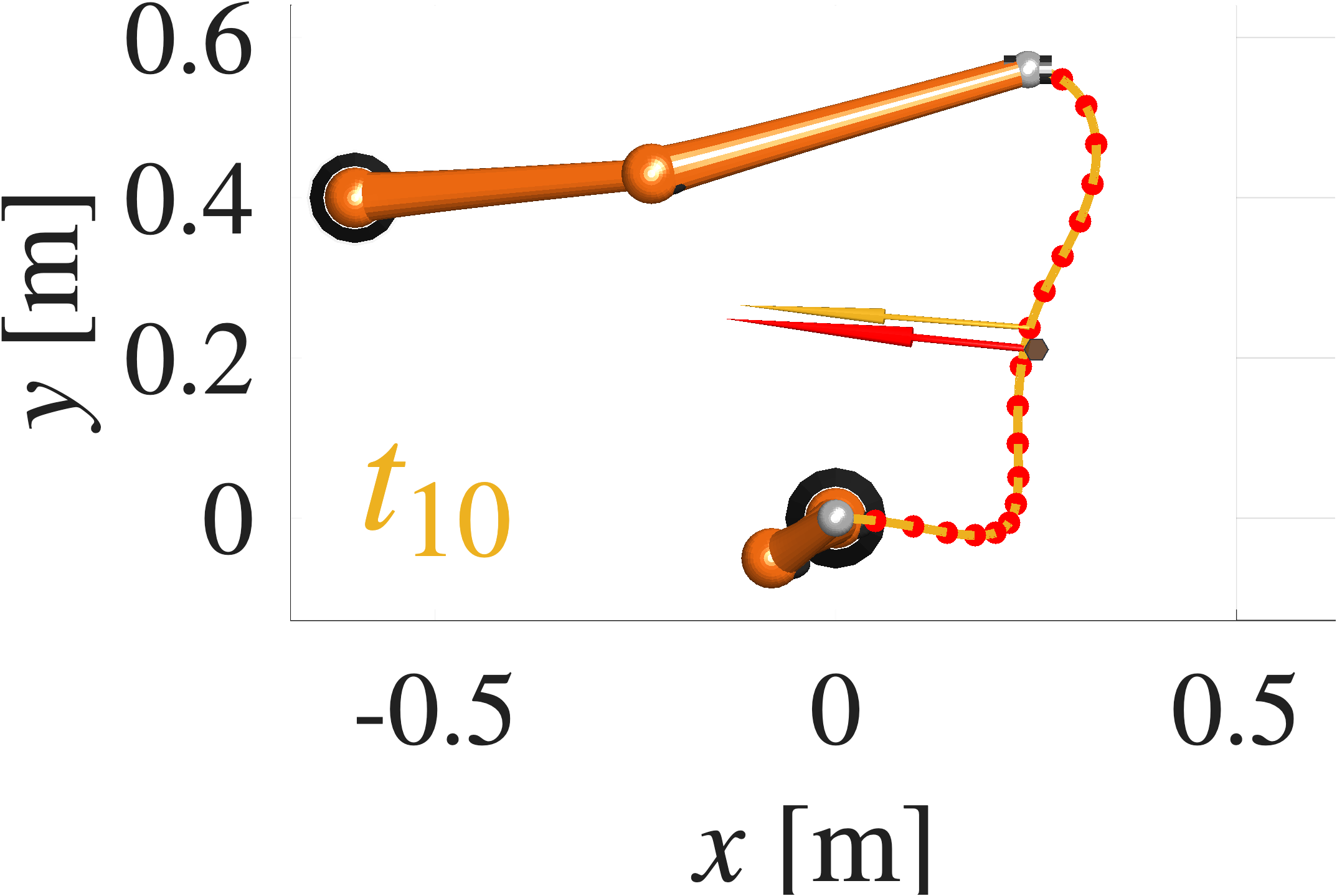}\par
        \vspace{0.5cm}
    \end{minipage}
    
    \includegraphics[width=0.98\linewidth]{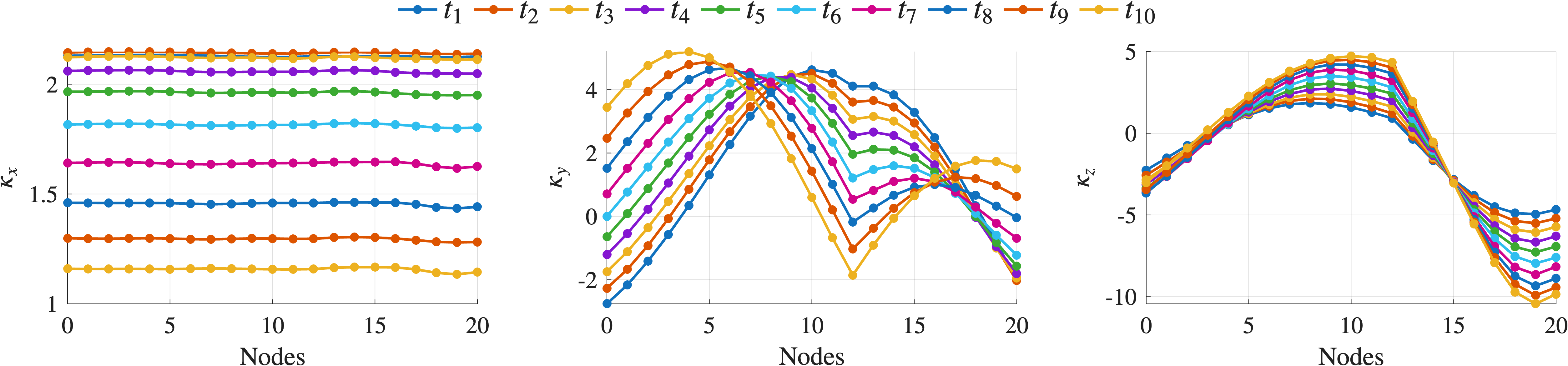}
    \caption{Intermediate configurations of the \textit{push} trajectory with a force acting at $s=0.6$ and markers measurement noise of $\sigma=1\times10^{-4}$~m. Left: 3D view at steps $t_1$--$t_{10}$. Right: top view at steps $t_2,t_4,t_6,t_8,t_{10}$. Bottom: estimated curvature components $\kappa_x$ (torsion), $\kappa_y, \kappa_z$ (bendings). Through the graph-based algorithm, node poses at discrete points along the rod and the external force are estimated and shown in red at each step.}
    \label{fig:push-traj-2}
\end{figure*}

\begin{table*}[tb]
    \centering
    \caption{Pull Trajectory -- Comparison Between True and Estimated Values}
    \label{tab:pull-traj-1}
    \footnotesize
    \setlength{\tabcolsep}{3pt}
    \begin{tabular*}{\textwidth}{@{\extracolsep{\fill}} c ccc ccc cc cc cc @{}}
    \toprule
    & \multicolumn{3}{c}{True Force [N]} & \multicolumn{3}{c}{Est. Force [N]} & True Loc. & Est. Loc. & \multicolumn{2}{c}{True Dir. [$^\circ$]} & \multicolumn{2}{c}{Est. Dir. [$^\circ$]} \\
    \cmidrule(lr){2-4} \cmidrule(lr){5-7} \cmidrule(lr){8-8} \cmidrule(lr){9-9} \cmidrule(lr){10-11} \cmidrule(lr){12-13}
    Step & $f_x$ & $f_y$ & $f_z$ & $\hat{f}_x$ & $\hat{f}_y$ & $\hat{f}_z$ & $s$ & $\hat{s}$ & $\alpha$ & $\varepsilon$ & $\hat{\alpha}$ & $\hat{\varepsilon}$ \\
    \midrule
    1 & 0.104 & 0.000 & 0.263 & 0.111 & -0.003 & 0.270 & 0.500 & 0.473 & -0.167 & 68.380 & -1.680 & 67.717 \\
    2 & 0.118 & 0.023 & 0.287 & 0.128 & 0.026 & 0.304 & 0.500 & 0.480 & 10.917 & 67.298 & 11.485 & 66.744 \\
    3 & 0.129 & 0.052 & 0.311 & 0.158 & 0.055 & 0.337 & 0.500 & 0.474 & 21.976 & 65.803 & 19.274 & 63.625 \\
    4 & 0.147 & 0.090 & 0.336 & 0.155 & 0.098 & 0.365 & 0.500 & 0.476 & 31.547 & 62.875 & 32.191 & 63.343 \\
    5 & 0.156 & 0.134 & 0.362 & 0.161 & 0.137 & 0.395 & 0.500 & 0.472 & 40.806 & 60.401 & 40.340 & 61.816 \\
    6 & 0.161 & 0.183 & 0.384 & 0.190 & 0.192 & 0.396 & 0.500 & 0.480 & 48.560 & 57.592 & 45.184 & 55.710 \\
    7 & 0.144 & 0.234 & 0.417 & 0.159 & 0.246 & 0.437 & 0.500 & 0.483 & 58.305 & 56.629 & 57.144 & 56.184 \\
    8 & 0.133 & 0.289 & 0.441 & 0.153 & 0.303 & 0.478 & 0.500 & 0.476 & 65.181 & 54.209 & 63.117 & 54.624 \\
    9 & 0.111 & 0.349 & 0.460 & 0.123 & 0.365 & 0.485 & 0.500 & 0.474 & 72.378 & 51.459 & 71.338 & 51.532 \\
    10 & 0.113 & 0.419 & 0.472 & 0.125 & 0.433 & 0.486 & 0.500 & 0.472 & 74.893 & 47.376 & 73.883 & 47.178 \\
    \midrule
    Mean err. & & & & 0.015 & 0.008 & 0.022 & & 0.024 & & & 1.454 & 0.829 \\
    \bottomrule
    \end{tabular*}
\end{table*}
~
\begin{table*}[tb]
    \centering
    \caption{Push Trajectory -- Comparison Between True and Estimated Values}
    \label{tab:push-traj-2}
    \footnotesize
    \setlength{\tabcolsep}{3pt}
    \begin{tabular*}{\textwidth}{@{\extracolsep{\fill}} c ccc ccc cc cc cc @{}}
    \toprule
    & \multicolumn{3}{c}{True Force [N]} & \multicolumn{3}{c}{Est. Force [N]} & True Loc. & Est. Loc. & \multicolumn{2}{c}{True Dir. [$^\circ$]} & \multicolumn{2}{c}{Est. Dir. [$^\circ$]} \\
    \cmidrule(lr){2-4} \cmidrule(lr){5-7} \cmidrule(lr){8-8} \cmidrule(lr){9-9} \cmidrule(lr){10-11} \cmidrule(lr){12-13}
    Step & $f_x$ & $f_y$ & $f_z$ & $\hat{f}_x$ & $\hat{f}_y$ & $\hat{f}_z$ & $s$ & $\hat{s}$ & $\alpha$ & $\varepsilon$ & $\hat{\alpha}$ & $\hat{\varepsilon}$ \\
    \midrule
    1 & -0.127 & -0.065 & -0.248 & -0.133 & -0.060 & -0.262 & 0.600 & 0.577 & -152.905 & -60.056 & -155.728 & -60.896 \\
    2 & -0.188 & -0.062 & -0.272 & -0.211 & -0.054 & -0.281 & 0.600 & 0.575 & -161.875 & -53.960 & -165.676 & -52.223 \\
    3 & -0.255 & -0.057 & -0.285 & -0.285 & -0.041 & -0.306 & 0.600 & 0.575 & -167.338 & -47.435 & -171.736 & -46.717 \\
    4 & -0.326 & -0.053 & -0.287 & -0.337 & -0.037 & -0.297 & 0.600 & 0.576 & -170.811 & -41.055 & -173.804 & -41.192 \\
    5 & -0.399 & -0.046 & -0.278 & -0.408 & -0.031 & -0.301 & 0.600 & 0.577 & -173.358 & -34.726 & -175.712 & -36.350 \\
    6 & -0.470 & -0.038 & -0.261 & -0.497 & -0.031 & -0.255 & 0.600 & 0.577 & -175.359 & -28.983 & -176.446 & -27.137 \\
    7 & -0.540 & -0.024 & -0.237 & -0.559 & 0.004 & -0.228 & 0.600 & 0.584 & -177.493 & -23.622 & 179.618 & -22.173 \\
    8 & -0.608 & -0.003 & -0.206 & -0.610 & 0.012 & -0.212 & 0.600 & 0.581 & -179.710 & -18.727 & 178.853 & -19.164 \\
    9 & -0.670 & 0.025 & -0.172 & -0.689 & 0.041 & -0.171 & 0.600 & 0.580 & 177.861 & -14.412 & 176.559 & -13.954 \\
    10 & -0.728 & 0.056 & -0.146 & -0.751 & 0.074 & -0.143 & 0.600 & 0.573 & 175.585 & -11.278 & 174.344 & -10.739 \\
    \midrule
    Mean err. & & & & 0.017 & 0.015 & 0.010 & & 0.022 & & & 2.432 & 0.979 \\
    \bottomrule
    \end{tabular*}
\end{table*}

\subsubsection{Evaluation Metrics}
The results are classified in terms of force magnitude, direction, and location accuracy. Tables~\ref{tab:pull-traj-1} and~\ref{tab:push-traj-2} therefore report the true quantities against the estimated ones. The force is given by its components $f_x, f_y, f_z$ in the world frame and its direction is expressed by the azimuth angle $\alpha = \operatorname{atan2}(f_y, f_x)$ and the elevation angle $\varepsilon = \arcsin(f_z/f_{ext})$, both in degrees, as illustrated in Figure~\ref{fig:az-el}. The location is the 
arc-length coordinate $s$ of the contact point. Estimated quantities are denoted by a hat.

\subsection{Trajectory Simulation}
The estimation is established under quasi-static conditions, where incremental variations in tip poses and external interaction forces yield a sequential trajectory of evolving rod shapes. This sequential nature enables the use of a warm-start strategy, where the optimization uses the last estimation's result as initial guess. To perform a comprehensive evaluation, two distinct trajectories with point contact were designed: a \textit{pull} trajectory, where the rod pulls an object by applying force from the inside of the rod outward, causing its initial U-shape to deform into a V-shape; a \textit{push} trajectory, where the rod is pushed against the object, generating a force from the outside inward, flattening the initial U-shaped configuration towards a M-shape.

It is worth noting that the force and tip pose values do not evolve continuously between successive shapes. Instead, they are updated discretely, with maximum steps of 0.05\,N for force, 62\,mm for tip position, and $6^\circ$ for tip orientation. This discrete updating represents a worst-case scenario where sensors' information is not continuously retained, but is available at discrete time steps.

Figures~\ref{fig:pull-traj-1} and~\ref{fig:push-traj-2} illustrate these trajectories across sequential steps, providing both 3D and planar-top views. In these plots, estimated rod shapes and force vectors are highlighted in red, while the baseline shapes obtained as in Sec.~\ref{sec:dataset-gen} have distinct colours across steps. 
As the figures show, the method reconstructs the force's magnitude, location, and orientation with high fidelity, even under severe structural deformation. Furthermore, the curvature profiles ($\kappa_x, \kappa_y, \kappa_z$) effectively capture the external stresses, exhibiting localized peaks precisely at the points of force application. Quantitative force estimation results are summarized in Tables~\ref{tab:pull-traj-1} and~\ref{tab:push-traj-2}.

In the \textit{pull} trajectory, the force is applied at the centre of the rod $s = 0.5$, producing symmetric deformation modes. The magnitude is estimated with mean errors $1.5\times10^{-2}$\,N, $8\times10^{-3}$\,N, $2.2\times10^{-2}$\,N for its $f_x,f_y,f_z$ components, respectively. Location is identified across all steps with a mean error of $2.4\times10^{-2}$m along the rod and direction is found with average azimuth and elevation errors of $1.5^\circ$ and $0.8^\circ$, respectively. 
Recalling that we also estimate the DLO shape, compared with the ground truth in Sec.~\ref{sec:dataset-gen}, across all the steps we achieve a mean position error per node of $1.6\times10^{-4}$\,m, with the minimum and the maximum errors of $1\times10^{-5}$\,m and $8.4\times10^{-4}$\,m, respectively. 

Conversely, the \textit{push} trajectory, represents a considerably more complex scenario. The contact location at $s=0.6$ breaks the symmetry of the shape 
and a significant axial force component introduces torsion. Despite these challenges, the estimation accuracy remains consistent, achieving mean magnitude errors of $1.7\times10^{-2}$\,N, $1.5\times10^{-2}$\,N, $1\times10^{-2}$\,N for $f_x,f_y,f_z$, respectively. The contact location is found with an average error of $2.2\times10^{-2}$m and azimuth and elevation mean errors are $2.4^\circ$ and $1^\circ$, respectively. In terms of shape error, compared with the ground truth in Sec.~\ref{sec:dataset-gen}, we achieve a mean position error per node of $1.6\times10^{-4}$\,mm, with the minimum and the maximum position error at a node reaching $1.3\times10^{-5}$\,mm and $5.1\times10^{-4}$\,mm, respectively.

Data show that the main estimation error is in terms of location $\hat{s}$ and they reveal an overestimation of the force magnitude especially in presence of torsion. Regarding the error on $\hat{s}$, this can be reduced by increasing $N$ while reducing the standard deviation $\sigma_f$ of the Gaussian that models the concentrated force. Here we select a trade-off between accuracy and computational cost. 
On the other hand, torsion increase the stiffness of the DLO thus requiring higher force to compensate the shape error. Notice that in these tests torsion could be estimated through tip poses only. Therefore, additional wrench sensors could be useful for torsion-dominant tasks.

\section{Conclusion}
\label{sec:conclusion}
In this paper, factor graphs were introduced to estimate the full contact force vector during dual-arm manipulation of DLOs. In this scenario, the lack of actuation and the strong coupling between external forces and internal wrenches lead to a challenging estimation procedure. Additionally, only one \textit{type} of sensor information was introduced in the estimator to stress the robustness of the method.

The approach was evaluated under quasi-static conditions on two distinct manipulation tasks, a symmetric \textit{pull} trajectory and a more challenging, asymmetric \textit{push} trajectory. In both cases, the 3D contact force was estimated without prior knowledge of its magnitude, direction or location, with markers corrupted by additive white Gaussian noise of standard deviation $\sigma = 1\times10^{-4}$\,m. 

Despite the challenges and relying only on one sensor \textit{type} -- markers positions from visual information -- the proposed method successfully recovers the contact force, its location and the DLO shape with minimal error even with non negligible torsion.

Future work will primarily focus on experimental validation of the proposed estimation technique. Subsequently, efforts will be spent towards the implementation of planning strategies, moving a step towards safe control and guidance.


\section{Data Availability}
The trajectory data will be available upon request.

\section{Acknowledgment}
U.S. Government sponsorship acknowledged. Dr. Quadrelli's contribution was carried out at the Jet Propulsion Laboratory, California Institute of Technology, under a contract with the National Aeronautics and Space Administration. Dr. Quadrelli was the mentor of Dr. Di Paola in the summer 2025 as part of the JPL Postdoctoral Visitor Program.

\bibliographystyle{IEEEtran}
\bibliography{references}

\end{document}